\documentclass[acmsmall,screen]{acmart}

\setcopyright{rightsretained}
\acmYear{2026}
\acmConference[ISSTA 2026]{ACM SIGSOFT International Symposium on Software Testing and Analysis}{October 3--9, 2026}{Oakland, CA, USA}

\usepackage{circledsteps}
\usepackage{mathtools}
\usepackage{xspace}
\usepackage{wrapfig}
\usepackage{algorithm}

\usepackage{algpseudocode}
\usepackage{subcaption}
\usepackage{multirow}
\usepackage{enumitem}
\usepackage{makecell}
\usepackage{arydshln}
\usepackage[normalem]{ulem}

\usepackage[capitalize,noabbrev]{cleveref}

\newcommand{\Tool}{ARQ\xspace}

\newcommand{\labels}{\mathcal{Y}}
\newcommand{\normal}{\mathcal{N}(0, \sigma^2 I)}
\newcommand{\palb}{\underline{p_A}}
\newcommand{\pbub}{\overline{p_B}}
\newcommand{\prob}{\mathbb{P}}
\newcommand{\eqlb}{\zeta_{x}}

\newcommand{\ACRorig}{ACR_{orig}}

\DeclareMathOperator*{\argmax}{arg\,max}

\newcommand{\XComment}[1]{}

\newcommand{\ArtifactURL}{\url{https://github.com/uiuc-arc/ARQ}}

\theoremstyle{plain}
\newtheorem{theorem}{Theorem}[section]

\theoremstyle{definition}

\theoremstyle{remark}

\begin{document}

\title{ARQ: A Mixed-Precision Quantization Framework for Accurate and Certifiably Robust DNNs}

\author{Yuchen Yang}
\affiliation{%
  \institution{University of Illinois Urbana-Champaign}
  \country{USA}
}
\email{yucheny8@illinois.edu}

\author{Yifan Zhao}
\affiliation{%
  \institution{University of Illinois Urbana-Champaign}
  \country{USA}
}
\email{yifanz16@illinois.edu}

\author{Shubham Ugare}
\affiliation{%
  \institution{University of Illinois Urbana-Champaign}
  \country{USA}
}
\email{sugare2@illinois.edu}

\author{Gagandeep Singh}
\affiliation{%
  \institution{University of Illinois Urbana-Champaign}
  \country{USA}
}
\email{ggnds@illinois.edu}

\author{Sasa Misailovic}
\affiliation{%
  \institution{University of Illinois Urbana-Champaign}
  \country{USA}
}
\email{misailo@illinois.edu}


\begin{abstract}
Mixed precision quantization has become an important technique for
optimizing the execution of deep neural networks (DNNs).
{\emph{Certified robustness}, which provides provable guarantees about a model's ability to withstand different adversarial perturbations, has rarely been addressed in quantization due to the unacceptably high cost of certifying robustness.} %
This paper introduces ARQ, an innovative mixed-precision quantization method that not only
preserves the clean accuracy of the smoothed classifiers, but also maintains
their certified robustness. ARQ uses reinforcement learning to find accurate and robust
DNN quantization, while efficiently leveraging randomized smoothing,
a popular class of statistical DNN verification algorithms.
\Tool{} consistently performs better than multiple state-of-the-art quantization techniques
across all the benchmarks and the input perturbation levels. The performance of ARQ quantized networks reaches that of the original DNN with floating-point weights, while using only $1.5\%$ instructions and the highest certified radius.
\Tool{}'s code is available at \ArtifactURL.
\end{abstract}


\maketitle

\section{Introduction}

Deep neural networks (DNNs) are increasingly deployed as components in safety-critical systems,
including autonomous driving, medical diagnosis,
and aviation~\citep{bojarski2016end,AMATO201347,acasxu:18}.
These safety-critical environments impose requirements on DNNs' accuracy, efficiency and reliability.
Prior works from both AI and SE communities have developed techniques to assess and improve
the reliability of DNNs and DNN-based systems~\citep{Ma_2018,Kim_2019,10.1145/3293882.3330579,riccio2020modelbasedexplorationfrontierbehaviours,neelofar2023reliableaiadequacymetrics}.
However, preserving DNN reliability becomes more challenging when efficiency constraints are introduced.

A defining quality of DNN reliability is its \emph{\textbf{robustness}}.
Robustness measures the ability of the DNN to produce correct outputs
when the input is perturbed under a certain threat model.
For example, classification DNNs should preserve the predicted classification label when the input is perturbed. Analyzing
\emph{\textbf{certified robustness}}  can provide guarantees on DNN robustness,
which combines \emph{robust verification} to formally define and measure the DNN's robustness,
and \emph{robust training} to improve robustness.
Certified robustness analyses systematically protect DNNs
against a broad scope of adversarial inputs~\citep{raghunathan2018certified,li2023sok}
but incur a high computational cost.
In particular, \emph{deterministic} robust verification techniques, such as DNN symbolic execution,
face significant scalability challenges when applied to large networks~\citep{sun2018concolictestingdeepneural,gopinath2020deepsafedatadrivenapproachchecking, 8987570}.
Many existing deterministic techniques have been used only on small networks and
datasets~\citep{singh2019boosting,lechner2022quantizationawareintervalboundpropagation,10.1145/3597926.3598127,10.1145/3540250.3558924,zhang22babattack}.
On the other hand, \emph{statistical} robustness verification methods bring better scalability.
\emph{Randomized smoothing} (RS) analyses~\citep{DBLP:conf/icml/CohenRK19} currently offer the greatest scalability,
scaling to ImageNet-sized models~\citep{li2021tsstransformationspecificsmoothingrobustness,yang2022certifiedrobustnessensemblemodels}.

As DNNs become larger and more compute-intensive, model compression techniques such as quantization
are gaining adoption to improve DNN inference efficiency.
Quantization reduces the model size and computational cost of DNNs,
improving inference speed and enabling large DNNs on memory-constrained devices.
Modern GPUs and custom accelerators~\cite{lowbit_fpga_1, lowbit_fpga_2,9689050,tahmasebi2025flexibitfullyflexibleprecision} increasingly support low-precision computation
to power quantized DNN inference.
Mixed-precision quantization allows different layers to be quantized with different bit-widths,
further improving compression rate while preserving accuracy.
However, quantizing a DNN reduces its robustness.
Recent empirical studies show that quantization can introduce behavioral discrepancies
between the original and quantized models,
which degrades robustness in deployment~\citep{hu2022characterizingunderstandingbehaviorquantized,Yahmed_2022,ijcai2019p800}.

Providing DNNs that are both efficient and robust poses a significant challenge.
Existing quantization strategies often maintain DNN accuracy after quantization,
but fail to consider DNN robustness.
RS and other certified robustness techniques have the potential to improve the robustness of quantized DNNs,
but due to their high computational overhead,
they have only been used as a post-quantization filter to reject low-robustness DNNs.
It remains an open research question to produce an accurate, robust and efficient DNN,
in one automated framework, that reduces the manual effort of programmers.

\vspace{0.05in}
\noindent{\bf Our Work: ARQ.} We present ARQ,
the first mixed-precision quantization framework that incorporates the analysis for certified robustness in the optimization objective and can handle state-of-the-art vision DNNs and datasets.
ARQ's algorithm takes a pre-trained DNN and %
automatically searches for \emph{\mbox{quantization} policies} -- the bit-widths
of the weights/activations of all layers in the DNN -- that (1)~preserve the DNN's
accuracy, (2)~improve its certified robustness, and (3)~reduce the model's computation cost and/or size.
This design allows ARQ to support mixed-precision quantization (MPQ), in which
 weights in each layer can be quantized with different bit-widths, thus
giving fine-grained \mbox{control over possible policies.}

The key insight behind ARQ is that the optimization for both
accuracy and robustness aims to maximize the DNN's \emph{certified radius},
which characterizes all the slightly perturbed inputs that the DNN classifies with the same label as
the non-perturbed input.
We use a common reinforcement learning (RL) framework to define the bitwidth search procedure. 
This optimization objective fits well within the RL framework and randomized smoothing is able to give probabilistic guarantees for state-of-the-art visual inputs (e.g., the size of ImageNet images).
Moreover, the design of ARQ's algorithm makes it possible to leverage recent approaches for incremental analysis of certified robustness~\cite{IRS} to further speed up the quantization policy search. 

We compare \Tool{} with existing searching and learning-based mixed-precision
quantizers. None of them can optimize for {certified robustness}. The baselines include
RL-based HAQ~\citep{wang2019haq}, integer programming-based LIMPQ~\citep
{tang2023mixedprecision}, differentiable algorithm NIPQ~\citep{shin2023nipqnoiseproxybasedintegrated}, Hessian-based HAWQ-V3~\citep{yao2021hawqv3dyadicneuralnetwork} and fixed-precision
quantization PACT~\citep{choi2018pact}. We evaluated four DNN
architectures commonly used in literature: ResNet-20 convolutional DNN and ViT-tiny vision transformer on CIFAR-10, and
ResNet-50 and {MobileNetV2 convolutional DNNs on ImageNet.}  

We demonstrate that \Tool{} consistently finds DNNs with higher certified robustness and clean accuracy than these baselines
across all the benchmarks and the input noise levels. In many cases, ARQ can reach, and sometimes even improve on the accuracy and robustness of the
original floating-point network, with only $1.5\%$ operations.

\vspace{0.05in}
\noindent{\bf Contributions.} The paper \mbox{makes several
contributions:}

\begin{itemize}[leftmargin=.1in]\itemsep 2pt
\item {\bf Approach:} We present ARQ, the first system for mixed-precision
 quantization that optimizes for certified robustness of DNNs. It poses an optimization problem that maximizes the certified radius for a bounded resource usage cost (e.g., compute instructions, model size). 

\item {\bf Analysis-Driven Framework:} \Tool{}'s algorithm incorporates randomized smoothing analysis within the reinforcement learning loop, which enables it to find certifiably robust \mbox{quantized DNNs.}

\item {\bf Results:} Our experiments on four commonly used networks show that \Tool{} finds DNNs with higher certified robustness and clean accuracy than the state-of-the-art quantization techniques.
\end{itemize}
Beyond its immediate contributions, \Tool{}'s design presents a blueprint for how other quantization systems can include certified robustness analyses (based on randomized smoothing or other deterministic methods) in their quantization search algorithms.

\section{Background}
\subsection{Mixed Precision Quantization}
\label{sec:mpq}

DNN quantization compresses the model to reduce the network's size and compute cost. %
Quantization applies to float-valued weights and activations in the network and converts them to integer values of certain bit-widths.
Using the same bit-width for the entire network is sub-optimal because some layers are more amenable to quantization than others.
Mixed Precision Quantization assigns different bit-widths
per weight or activation in a network
and searches for the best combination of bit-widths.
A \textit{quantization policy} \(P\) is a sequence of bit-width assignments
to each layer in the network.
For a network of $L$ layers, where each layer has
$N$ bit-width options $\{b_1, b_2, \dots, b_N\}$ for both weights and activations,
there are $N^{2L}$ combinations of quantization policies.
We can then formulate the process of \mbox{optimizing} the quantization policy $P$ for a network
$f_P(x)$ 
as the following mathematical optimization problem:
\begin{equation}
P_{optimal} = \arg\max_{P \in \mathcal{P}} \text{Acc}(f_P(x), y) \quad \text{s.t. } \text{Cost}(f_P) < C_0
\label{eq:mpq}
\end{equation}
\begin{equation}
\text{Acc}(f(x), y) = \frac{1}{|X|} \sum_{(x, y) \in X} \mathbf{1}(f(x) = y)
\label{eq:acc}
\end{equation}
$\mathcal{P}$ is the set of all quantization policies and
$P_{optimal} \in \mathcal{P}$ is the optimal policy that maximizes the accuracy of the quantized DNN $\text{Acc}(f_P(x))$ on dataset $X$.
$\text{Cost}(f_P)$ is the resource usage of the quantized DNN
(e.g., the model size, the number of compute bit operations, or energy consumption),
and $C_0$ is a user-specified bound on the resource. By sweeping over multiple values of $C_0$, one can find a Pareto-optimal set of policies that maximize \mbox{both accuracy and cost.}

\noindent\textbf{Reinforcement Learning Based Quantization.} Researchers~\cite{wang2019haq, lou2020autoq}
applied Reinforcement Learning (RL) to the search for quantization policies.
One of the popular RL algorithms in this context is the Deep Deterministic Policy Gradient (DDPG)~\mbox{\citep{lillicrap2019continuouscontroldeepreinforcement}}. 
The DDPG agent iteratively interacts with the environment (the neural network) by observing the state $S_k$ (the configuration of the $k_{th}$ layer), taking an action $a_k$ (the quantization bit-width), and receiving a \mbox{reward $r$ (the resulting accuracy).}

\subsection{Certified Neural Network Robustness Analysis}
\label{sec:cnr}
A classifier is \textit{certifiably robust}
when its predictions are guaranteed to remain consistent within a neighborhood of input $x$.
Consider a classification problem from $\mathbb{R}^m$ to classes $\labels$.
Let $f: \mathbb{R}^m \to \labels$ be a neural network classifier.
We seek a \textit{smoothed} classifier $g: \mathbb{R}^m \to \labels$
whose prediction matches $f$ for any input $x$
and is \textit{constant} within some neighborhood of $x$, which means $g$ is \textit{certifiably robust}.
Randomized smoothing \citep{DBLP:conf/icml/CohenRK19,yang2020randomized,NEURIPS2020_1896a3bf} provides a way to construct
such a smoothed classifier $g$ from the base classifier $f$.
When queried at $x$, $g$ returns the class that
$f$ is most likely to return when Gaussian noise is added to $x$:
\begin{equation}
    g(x) := \argmax_{y \in \labels} \prob(f(x+\varepsilon) = y) \  \text{s.t.} \  \varepsilon \sim \normal \label{eq:gx}
\end{equation}
\citet{DBLP:conf/icml/CohenRK19}
show that $g$'s prediction
is constant within an $l_2$ ball around any input $x$.
The radius of that ball, $R(x)$, is known as the \textit{certified radius}. 
$\varepsilon$ is the Gaussian noise used in the smoothing process (Eq.~\ref{eq:gx}),
sampled from Gaussian distribution of mean $0$ and variance $\sigma^2 I$
($I$ is the identity matrix).
$\sigma$ is the noise level, a hyperparameter of the smoothed classifier $g$
independent of the input $x$.
The \textit{certified accuracy} of a classifier is defined as the probability that the classifier correctly predicts the true labels of samples $x$ for which the certified radius $R(x)$ exceeds a certain threshold~$r$.
The \mbox{\textit{clean accuracy}} is the certified accuracy when $r = 0$.

\begin{theorem}[From \cite{DBLP:conf/icml/CohenRK19}]\label{thm:rs}
Suppose $c_A \in \labels$, $\palb, \pbub \in [0, 1]$. If
\begin{equation*}
    \prob (f(x+\epsilon) = y_A) \geq \palb \geq \pbub \geq \max_{y \neq y_A} \prob (f(x+\epsilon) = y),
    \label{eq:rs}
\end{equation*}
\noindent{}then $g(x+\delta) = y_A$ for all adversarial perturbations $\delta$ satisfying $\|\delta\|_2 \leq \frac{\sigma}{2} (\Phi^{-1}(\palb) - \Phi^{-1}(\pbub))$, where $\Phi^{-1}$ denotes the inverse of the standard Gaussian CDF.
\end{theorem}
Computing the exact probabilities $\palb, \pbub$ from Thm.~\ref{thm:rs}
is intractable in general. 
For practical applications, RS certification utilizes sampling to estimate 
$\palb$ and $\pbub$ using the Clopper-Pearson method \citep{10.2307/2331986}.
If using this procedure yields $\palb > 0.5$, then the RS algorithm sets $\pbub = 1 - \palb$ and computes the certified radius as
\begin{equation}
R(x) = \sigma\cdot\Phi^{-1}(\palb)    
\end{equation}
via Theorem~\ref{thm:rs}; otherwise, it returns ABSTAIN, i.e., it cannot prove the certified robustness.
This means $g(x+\delta) = y_A$ for all adversarial perturbations $\delta$ satisfying $\|\delta\|_2 \leq R(x)$.

\begin{wrapfigure}{r}{0.5\textwidth}
  \centering
  \vspace{-0.1in}
  \includegraphics[width=0.48\textwidth]{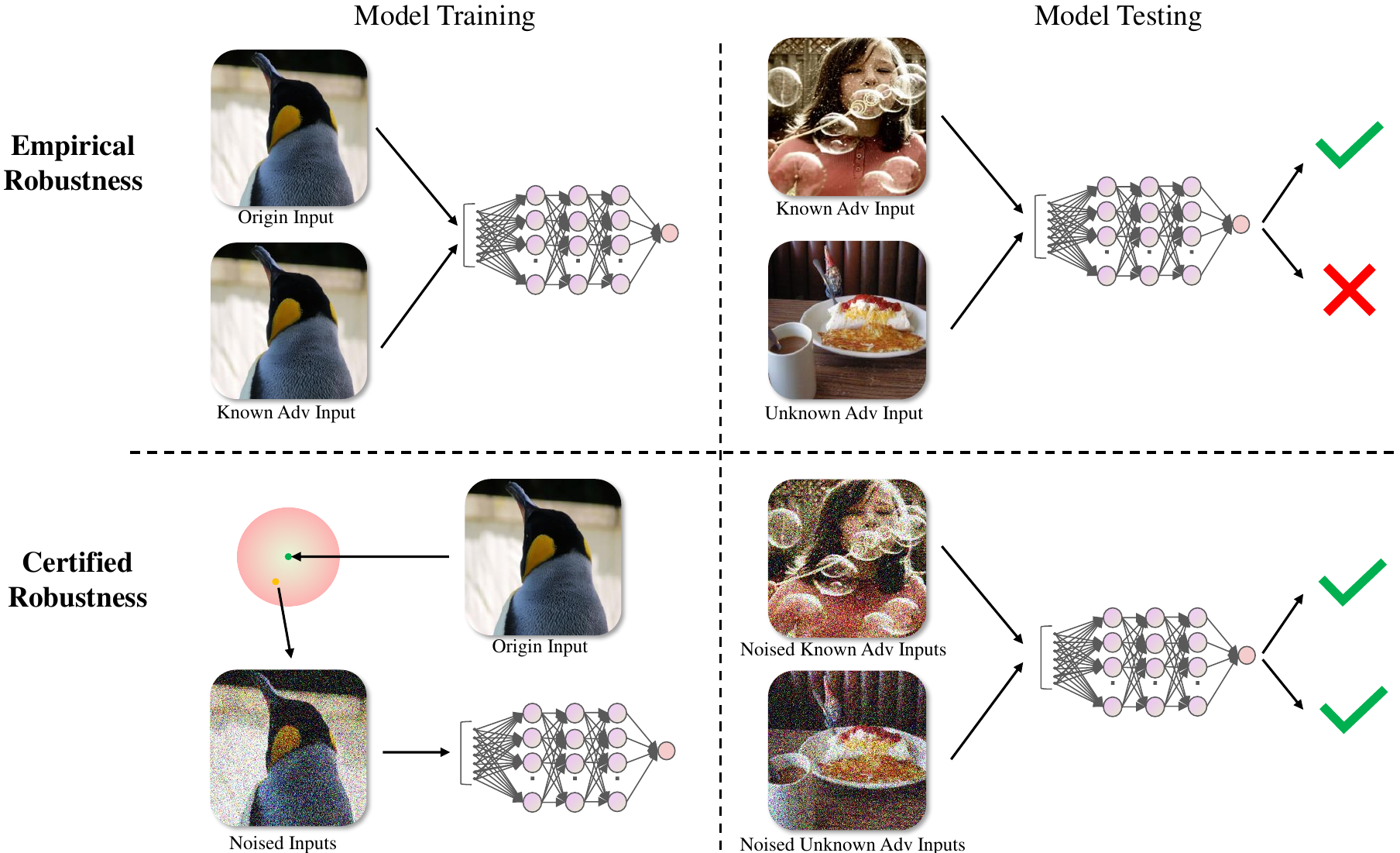}
  \caption{Difference between empirical and certified robustness. Images from ImageNet fed into ResNet-50. Known attacks use FGSM~\citep{DBLP:journals/corr/GoodfellowSS14}, unknown attacks use PGD~\citep{madry2019deeplearningmodelsresistant}. Empirical defense: Pixel Deflection~\citep{prakash2018deflectingadversarialattackspixel}.}
  \label{fig:robust}
\end{wrapfigure}

\noindent\textbf{Certified vs.\ Empirical Robustness.}
\label{app:robust}
Figure \ref{fig:robust} compares empirical robustness and certified robustness.
Empirical robustness methods rely on known adversarial attacks during both training and testing. Although effective against specific attack strategies, they do not provide guarantees against future or unseen adversarial methods.
Certified robustness methods ensure that the model's predictions remain unchanged within a mathematically guaranteed region around the input. By certifying robustness during both training and testing, these methods offer provable guarantees against all bounded perturbations, including those from unknown adversarial methods.

\section{\Tool{} Approach}
\subsection{Problem Statement}\label{sec:problem}
\Tool{} provides a mixed-precision quantization method
that optimizes both the robustness and accuracy of the \textit{quantized smoothed classifier} $g_P$
from a base classifier~$f$.

\noindent\textbf{Quantization Challenges.}
To improve the robustness of the quantized smoothed classifier $g_P$,
one could naively replace the accuracy metric in the formulation of an existing MPQ method (Eqn.~\ref{eq:mpq})
with an accuracy metric for the base classifier $f_P$ that uses Gaussian-noised inputs. This objective seeks to maximize the accuracy of the base classifier $f_P$ over the data distribution $X$ with the Gaussian noise $\varepsilon \sim \mathcal{N}(0, \sigma^2 I)$:
\begin{equation}
    P_{optimal} = \arg\max_{P \in \mathcal{P}} \text{Acc}(f_P(x+\varepsilon), y) \quad \text{s.t. } \quad \text{Cost}(f_P) < C_0
\label{eq:naive_baseline}
\end{equation}
\begin{itemize}[leftmargin=.1in]\itemsep 1pt
\item By taking one Gaussian-noised sample per data point, 
this accuracy metric is highly affected by randomness. It fails to accurately estimate the probability $\underline{\prob(f_P(x+\varepsilon) = y)}$ over the noise distribution, and as a result, does not properly capture the robustness of $g_P$.

\item The accuracy of the quantized base classifier $f_P$ correlates directly with the average lower bound probability ($\palb$) as stated in Eqn.~\ref{eq:rs}. However, improving the quantized base classifier $f_P$'s accuracy on samples with $ \palb < 0.5 $ does not improve the quantized smoothed classifier $g_P$'s accuracy. For samples where $ \palb < 0.5 $, the certified radius is less than zero, indicating that the quantized smoothed classifier $g_P$ cannot provide any robustness guarantee for these samples. Consequently, these samples cannot be correctly classified by the quantized smoothed classifier $g_P$, regardless of the improvements made to the quantized base classifiers on such inputs.

\item Using accuracy as the optimization goal does not efficiently reflect the robustness of the neural networks. 
As $R = \sigma\cdot\Phi^{-1}(\palb)$, the certified radius $R$ is related to the probability $\palb$ via the inverse normal CDF, $\Phi^{-1}$.
This function is convex for $\palb \in [0.5, 1)$, meaning its slope increases rapidly as $\palb$ approaches 1. Therefore, optimizing only for the \mbox{accuracy} of the quantized base classifier $f_P$ disproportionately prioritizes small, easy gains in $\palb$ on low-radius samples, while neglecting the much larger radius improvements that come from difficult gains in $\palb$ on high-radius samples.
\end{itemize}

\paragraph{ARQ Optimization Objective.} Instead, we propose using the certified radius of smoothed classifiers to directly guide the quantization method. It is more explicit as it uses feedback from the smoothed classifiers instead of the base classifiers, and it can combine the goal of optimizing both the clean accuracy and the robustness of the smoothed classifiers. 
We define the following optimization problem to find the optimal quantization policy: %
\begin{equation}
P_{optimal} = \arg \max_{P \in \mathcal{P}} ( \text{Average Certified Radius} ) \quad \text{s.t. }\quad \text{Cost}(f_P) < C_0
\end{equation}
where Average Certified Radius (ACR) is estimated as:  %
\begin{equation}
\text{ACR} = \frac{\sigma}{{|X|}}\sum_{(x, y) \in X} \Phi^{-1}(\underline{\prob(f(x+\varepsilon) = y)}), \quad \varepsilon \sim \normal \  \text{for each input} \ x
\label{eq:ACR}
\end{equation}
The $\underline{\prob(f_P(x+\varepsilon) = y)}$ here represents the lower bound of probability that base classifier $f$ can correctly classify input $x$ under noise $\varepsilon$. This follows from the definition of the certified radius $R(x) = \sigma\cdot\Phi^{-1}(\palb)$ for a given input $x$ and $\prob (f(x+\epsilon) = y_A) \geq \palb$ in Section \ref{sec:cnr}. By averaging over inputs in the dataset $X$, we obtain the ACR, which provides a measure of the overall robustness of the classifier. Since the clean accuracy of smoothed classifiers is the percentage of samples with a certified radius greater than zero, by focusing on optimizing the Average Certified Radius, we can improve both the accuracy and robustness of the quantized smoothed classifiers.

Therefore, our final robustness-aware quantization problem formulation is: 
\begin{equation}
    \boxed{
        P_{optimal} = \arg\max_{P \in \mathcal{P}} \sum_{(x, y) \in X} 
        \Phi^{-1}(\underline{\prob(f_P(x+\varepsilon) = y)}) \quad \text{s.t. } \quad \text{Cost}(f_P) < C_0
    }
    \label{eq:ACR_all}
\end{equation}

However, it is challenging to use this formulation to search exhaustively across quantization policies because calculating the certified radius, specifically obtaining $\underline{\prob(f_P(x+\varepsilon) = y)}$, is expensive -- this probability is estimated using the Clopper-Pearson method \citep{10.2307/2331986}, and the confidence level is related to the number of samples, and may require thousands of samples even for a single image~\citep{DBLP:conf/icml/CohenRK19}.
Instead, we employ a reinforcement learning agent to search for the optimal quantization policy $P_{optimal}$, which we describe next.

\subsection{\Tool{} Search Algorithm}
\label{sec:alg}
\Tool{}'s search algorithm (Algo.~\ref{alg:arq}) %
determines the optimal quantization policy for a given DNN~$f$.
We first {fully} certify the robustness of $g$, the smoothed version of $f$ {using a large number of samples} $n_0$ with function FullRobustCertify, and store the average certified radius of $g$ as $\ACRorig$ (line \ref{alg:arq:smoothorigin}, \ref{alg:arq:certifyorigin}).

During each episode $t$, we first initialize the list A to store the actions across all layers. And on each layer, our RL agent first observes the $k_{th}$ layer's configuration as state $S_k$ and uses the policy network $\mu(\cdot)$ learned from the previous episodes to determine an action $a_k$ (line \ref{alg:arq:lineAction}).
For each layer, the agent selects two actions for the weights and the activations of that layer. The transition $(S_k, a_k, Reward, S_{k+1})$ is stored in the replay buffer $D$ for training the agent's policy network (line \ref{alg:arq:lineA}). Here, $Reward$ is only updated when the actions for all layers are determined. $S_{k+1}$ is the configuration of the next layer.

\begin{figure}[t]
\begin{algorithm}[H]
\small
\caption{\Tool{} Search Algorithm}
\label{alg:arq}
\textbf{Inputs:} $\Circled{f\,}$: original DNN, $\Circled{\sigma}$: standard deviation, $\Circled{X}$: inputs to the DNN, $\Circled{n_0}$: number of Gaussian samples used for original certification, $\Circled{n}$: number of Gaussian samples used for quantized model certification, $\Circled{C_0}$: constraint bound on the quantized models, $\Circled{N}$: the number of episodes for search, $\Circled{L}$: the number of layers, $\Circled{D}$: empty replay buffer, $\Circled{\,\mu(\cdot)\,}$: the policy network of the agent.
\begin{algorithmic}[1]
\Function{Policy Search}{$f, \sigma, X, n_0, n, C_0, N, L, D, \mu(\cdot)$}
\State $g \gets \text{SmoothedClassifier}(f, n_0)$\label{alg:arq:smoothorigin}; \State $\ACRorig \gets \text{FullRobustCertify}(g, X, \sigma)$\label{alg:arq:certifyorigin} \Comment{Certify original model}
\State $P_{optimal} \gets \varnothing$;\ $g_{P_{optimal}} \gets \varnothing$ \Comment{Initialize tracking variables}
\State $Reward_{best} \gets -\infty$;  \ $Reward \gets -\infty$
    \For{$t = 1$ \textbf{to} $N$} \Comment{Outer loop: traverse episodes}
        \State $A \gets \varnothing$ \Comment{Initialize action list for this episode}
        \For{$k = 1$ \textbf{to} $L$} \Comment{Inner loop: traverse layers}
            \State Observe the $k_{th}$ layer's state $S_k$
            \State Select action $a_k$ using policy $\mu(S_k)$ with exploration noise\label{alg:arq:lineAction}
            \State Observe next layer's state $S_{k+1}$
            \State Store transition $(S_k,\,a_k,\,Reward,\,S_{k+1})$ in replay buffer $D$ \label{alg:arq:lineA}
            \State Add $a_k$ to list $A$ \Comment{Collect $a_k$ for the full model}
        \EndFor
        \State $P_{t} \gets \text{CombineActionsToPolicy}(A, C_0)$ \label{alg:arq:linecombine} \Comment{Build policy from actions}
        \State $f_P \gets \text{Quantize}(f, P_t)$; \  \Comment{Apply quantization}
        \State $f_P \gets \text{FineTune}(f_P, X, \sigma)$\label{alg:arq:quantize} \Comment{Fine-tune quantized model}
        \State $g_P \gets \text{SmoothedClassifier}(f_P, n)$\label{alg:arq:smooth} \Comment{Certify quantized model}
        \State $ACR_P \gets \text{IncrementalRobustCertify}(g_P, X, \sigma)$\label{alg:arq:certify}
        \State $Reward_t \gets ACR_P-\ACRorig$\label{alg:arq:reward} \Comment{Compute reward}
        \If{$Reward_t > Reward_{best}$} \Comment{Update best if improved}
            \State $Reward_{best} \gets Reward_t$
            \State $P_{optimal} \gets P_t$; \ $g_{P_{optimal}} \gets g_P$
        \EndIf
        \State The $Reward$ for all transitions in this iteration is set to final $Reward_t$\label{alg:arq:set}
        \State Update Q-function, policy and target network. Reset the state.
    \EndFor
    \State \Return $(P_{optimal}$,\, $g_{P_{optimal}})$ \Comment{Return optimal policy}
\EndFunction
\end{algorithmic}
\end{algorithm}
\vspace{-.3in}
\end{figure}

After the RL agent proposes the actions for all layers, we first transform the continuous actions in list $A$ into discrete bit-widths and combine them into a quantization policy list $P_t$ using the function CombineActionsToPolicy. 
Then we evaluate the resource usage of the base classifier $f_P$, which is quantized through $P_t$. 
If the proposed quantization policy $P_t$ exceeds the specified resource constraint $C_0$, we will sequentially decrease the bit-width of each layer until the constraint is finally satisfied (line \ref{alg:arq:linecombine}).

The function $\text{Quantize}(f, P_t)$ represents the quantization on $f$ to $f_P$ with quantization policy $P_t$, where the floating-point weights and activations are mapped to integers.
We finetune $f_P$ for one epoch using a subset of inputs in dataset $X$ with Gaussian noise of size $\sigma$ to help it recover performance (line \ref{alg:arq:quantize}). 
{Line~\ref{alg:arq:smooth} smooths $f_P$ into a quantized smoothed classifier $g_P$ with $n$ (which is $ \ll n_0 $) samples.
Function IncrementalRobustCertify certifies the robustness of $g_P$ incrementally by reusing the information from the initial certification of $g$ to obtain the ACR of $g_P$ as $ACR_P$. 
}
The RL agent's reward for all actions, $Reward_t$, is set as $ACR_P - \ACRorig$, using the average certified radius of $g_P$ and $g$ to guide the learning of the agent (line \ref{alg:arq:reward}, \ref{alg:arq:set}).
After $N$ episodes, we obtain the optimal quantization policy $P_{\text{optimal}}$, with maximum average certified radius of the quantized \mbox{smoothed classifier.}

Following previous work \citep{He_2018, wang2019haq}, we use DDPG as our RL agent to search the bit-widths. 
At the $k_\text{th}$ layer of the base classifier $f$, the state $S_k$ of the agent is:
\begin{equation}
S_k = \left(
\begin{aligned}
&k, c_{\text{in}}, c_{\text{out}}, s_{\text{kernel}}, s_{\text{stride}}, \\
&s_{\text{feat}}, n_{\text{params}}, i_{d}, i_{wa}, a_{k-1}
\end{aligned}
\right)
\end{equation} 
where $c_{\text{in}}$ and $c_{\text{out}}$ are input/output channels, $s_{\text{kernel}}$, $s_{\text{stride}}$ and $s_{\text{feat}}$ are kernel, stride and feature map sizes, $n_{\text{params}}$ is the parameter count, $i_{d}$ and $i_{wa}$ indicate depthwise layers and 
weights/activations, and $a_{k-1}$ is the previous layer's action. The first and last layers are quantized at \mbox{8-bit.}

We use a continuous action space with $a_{\text{min}}=0$ and {$a_{\text{max}}=1$} to keep the relative order information among different actions. Observing the state $S_k$ and using the policy network $\mu(\cdot)$, the action $a_k$ is selected for the $k_{th}$ layer.
We then round $a_k$ into discrete bit-width $b_k$:
\begin{equation}
b_k = \text{round}(b_{\text{min}} - 0.5 + a_k \times (b_{\text{max}} - b_{\text{min}} + 1))
\label{eq:action}
\end{equation}
with $b_{\text{min}}$ and $b_{\text{max}}$ here denoting the min and max bit-width.

As lines \ref{alg:arq:lineA} and~\ref{alg:arq:linecombine} in Alg.~\ref{alg:arq} state, the actions %
are combined into a list 
$A = (a_1, a_2, \ldots, a_k, a_{k+1}, \ldots, a_L)$
where $a_L$ is the final action the agent made for the last layer.
When all layers have been traversed by the agent, the list $A$ is transformed into a discrete bit-width policy $P_t$ for the $t_{th}$ iteration:
$P_t = (b_1, b_2, \ldots, b_k, b_{k+1}, \ldots, b_L)$
which is also limited by the resource constraint $C_0$. When the given $P_t$'s resource usage exceeds $C_0$, the bit-width will be decreased sequentially from back to front.

Due to the high cost of certifying the robustness of the quantized smoothed classifier $g_P$, we perform the certification only at the end of an episode, which indicates all actions have been taken and the entire quantization policy $P_t$ comes out. This avoids the frequent and expensive robustness certification for each individual quantized layer. We define our reward function $Reward_t$ to be related to only the average certified radius of the smoothed classifiers (line \ref{alg:arq:reward}):
\begin{equation}
Reward_t = ACR_P - \ACRorig
\label{eq:reward}
\end{equation}
where $ACR_P$ denotes the average certified radius gained by the quantized smoothed classifier $g_P$ through the current quantization policy $P_t$, and $\ACRorig$ denotes the average certified radius of the original smoothed classifier $g$, which is a function of $x$ as formulated in Eqn.~\ref{eq:ACR}.

We find the optimal quantization policy $P_{optimal}$ for which the corresponding $g_{P_{optimal}}$ achieves the highest ACR by comparing $Reward_{best}$ with $Reward_t$.

The experiences in the form of transitions $(S_k,\,a_k,\,Reward_t,\,S_{k+1})$ are stored in the replay buffer $D$ to update the Q-function, policy, and target network of the DDPG agent (line \ref{alg:arq:set}).

Since our optimization goal is to maximize the average certified radius of $g_P$ across the entire policy $P_t$, we set the reward for all actions across different layers in one iteration to be the same value: the final reward $Reward_t$. This ensures the reward reflects the overall effectiveness of the quantization policy rather than individual layer actions, \mbox{promoting a more comprehensive optimization.}

\noindent{\bf Quantization.} We use a linear quantization method, which maps the floating-point value to discrete integer values in the range $[-c, c]$ for weights and $[0, c]$ for activations. The quantization function $\text{Quantize}(\cdot)$ that quantizes floating-point weight value $v$ to $b$-bit integer value $q$ can \mbox{be expressed as:}
\begin{equation}
q = \text{round}(\text{clip}({v} / {s}, -c, c)) \times s
\label{eq:quantize}
\end{equation}
where $v$ is the floating-point value, and $q$ is the quantized value.
$s = \frac{c}{2^{b-1}-1}$ is the scaling factor. 
$c$ is optimized through the KL-divergence between $q$ and $v$. 
In the network, each layer utilizes two distinct $c$ values for quantizing weights and activations.

\noindent{\bf IRS for Speedup in Certification.} Due to the significant time consumption caused by both the number of iterations required for quantization policy search and the time-consuming process of certifying each iteration of the quantized neural networks, we employ Incremental Randomized Smoothing (IRS) \citep{IRS} to certify quantized smoothed classifiers more efficiently. It is known that IRS can have similar precision to re-running RS when verifying networks that have sufficient structural similarity. Our design of \Tool{}'s algorithm aims to promote this property.

\section{Experimental Methodology}
\label{sec:methodology}

\begin{figure*}[b]
  \centering
  \begin{subfigure}[b]{0.45\textwidth}
    \centering
    \includegraphics[width=.85\textwidth]{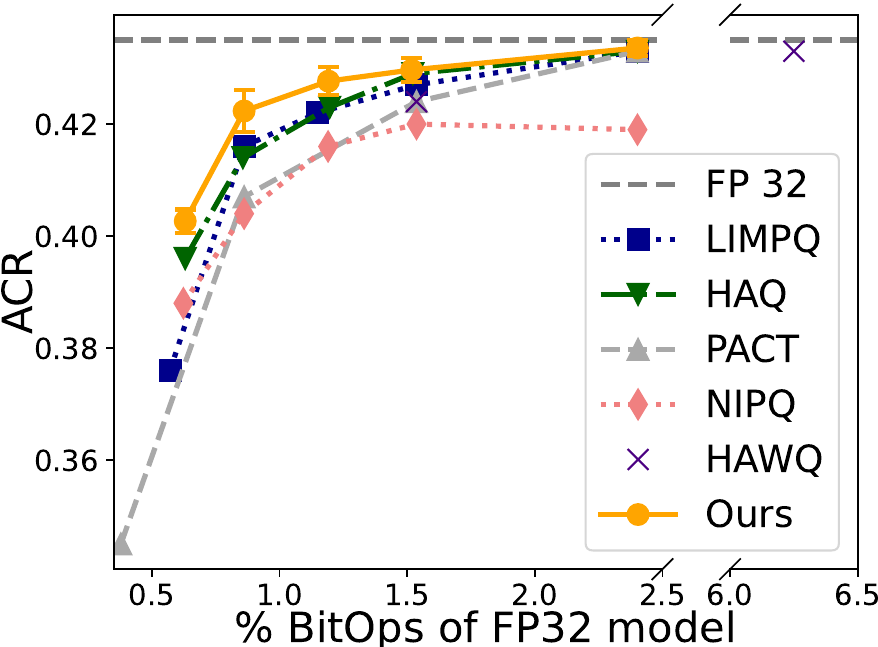}
    \caption{$\sigma = 0.25$}
    \label{fig:1a}
  \end{subfigure}
  \hfill
  \begin{subfigure}[b]{0.45\textwidth}
    \centering
    \includegraphics[width=.85\textwidth]{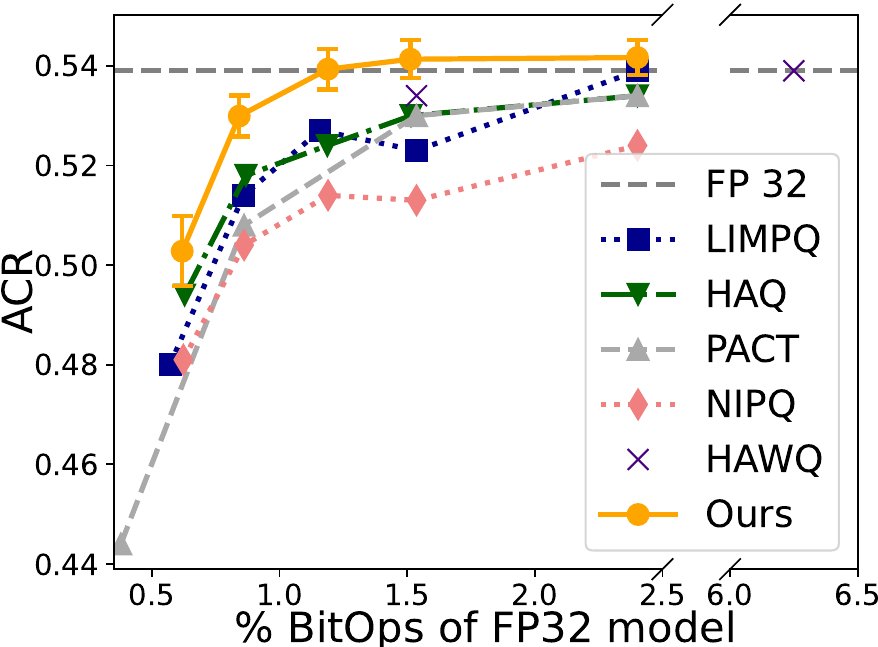}
    \caption{$\sigma = 0.5$}
    \label{fig:1b}
  \end{subfigure}
  
  \vspace{0.5em}
  
  \begin{subfigure}[b]{0.45\textwidth}
    \centering
    \includegraphics[width=.85\textwidth]{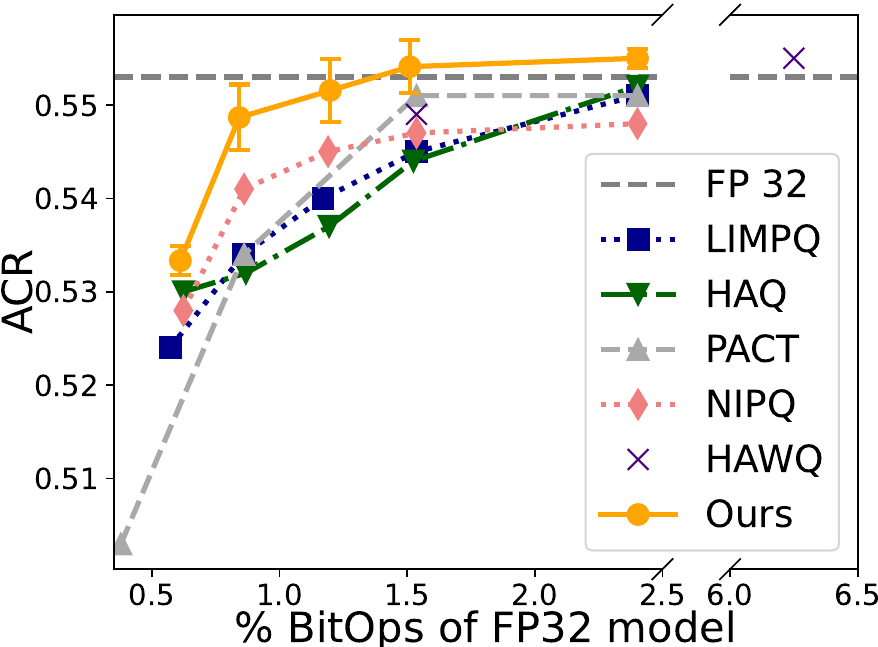}
    \caption{$\sigma = 1.0$}
    \label{fig:1c}
  \end{subfigure}
  \hfill
  \begin{subfigure}[b]{0.45\textwidth}
    \centering
    \includegraphics[width=.85\textwidth]{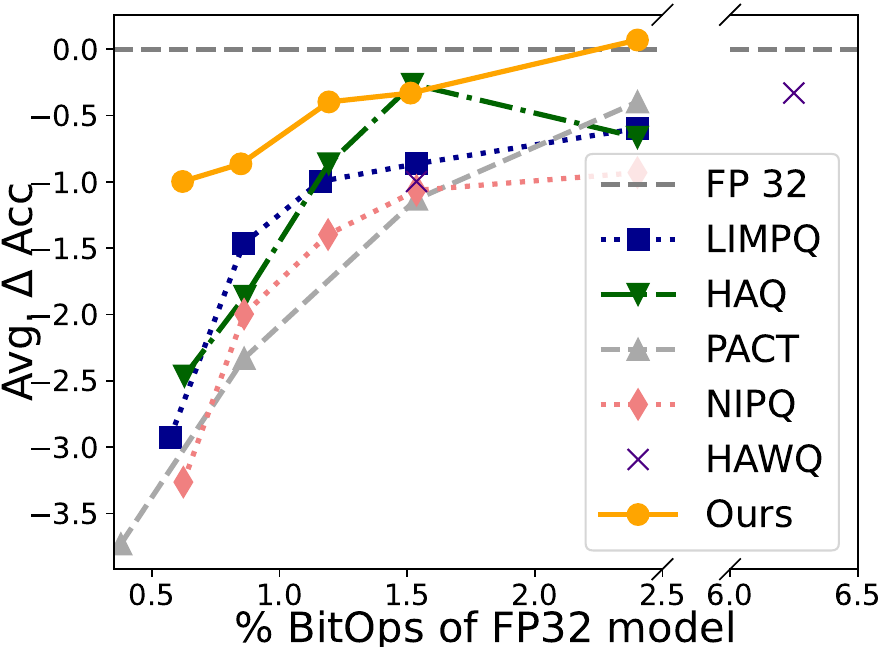}
    \caption{Avg. $\Delta$ Clean Acc}
    \label{fig:1d}
  \end{subfigure}
    \caption{Experiments on CIFAR-10. The x-axis shows the percentage of BitOPs of $f_P$ relative to the original floating-point $f$. The y-axis shows the ACR for the first three subfigures, and the average difference in clean accuracy between the methods and the original floating-point network across different $\sigma$ settings for Figure \ref{fig:1d}. Error bars denote 1-sigma standard deviation over 3 random seeds.}
  \vspace{-.2in}
  \label{fig:1}
\end{figure*}

\noindent{\bf Networks and Datasets.} 
We evaluate \Tool{} on CIFAR-10 \citep{cifar10} and ImageNet \citep{deng2009imagenet} datasets. 
We conduct all experiments on CIFAR-10 and ImageNet with 4-bit equivalent quantization.
We also perform ablation studies on CIFAR-10 with various quantization levels, as illustrated in Figure~\ref{fig:1}.
All models are trained and evaluated with Gaussian noise of variance $\sigma^2 \in \{0.25, 0.5, 1.0\}$ on inputs. 
We use ResNet-20 as the base classifier for CIFAR-10, and ResNet-50 and MobileNetV2 for ImageNet.
These models are chosen because they are widely used classifiers in previous studies within the areas of quantization and robustness.

\noindent{\bf Metrics.} We use the number of bit operations (BOPs, BitOPs) constraint for all methods following \citet{yao2021hawqv3dyadicneuralnetwork}. 
For example, the BitOPs of a convolution layer is
$\text{BOPs}(k) = {b_w b_a \cdot |K| w h} / {s^2}$
where $b_w$ and $b_a$ are the bitwidths for weights and activations,
$|K|$ is the number of parameters in the convolution filter,
$w$, $h$ are the width and height of the input, and $s$ is the stride of the convolution.

\noindent{\bf Robustness Certification.}
For the evaluation, we use confidence parameters $\alpha=0.001$ for the certification of the original smoothed classifier $g$,
following the setting used by RS \citep{DBLP:conf/icml/CohenRK19} and IRS \citep{IRS}. For policy search, we use 500 validation images, $n_0=10000$, and $n=500$ samples/image. For $\eqlb$ estimation, we use $\alpha=0.001$ and $\alpha_\zeta=0.001$ on CIFAR-10, and $\alpha=0.01$ and $\alpha_\zeta=0.01$ on ImageNet.

\noindent{\bf Evaluation and Baseline Comparison.}
We compare \Tool{} with state-of-the-art searching and learning-based mixed-precision quantization methods HAQ \citep{wang2019haq}, LIMPQ \citep{tang2023mixedprecision}, NIPQ \citep{shin2023nipqnoiseproxybasedintegrated} and HAWQ-V3 \citep{yao2021hawqv3dyadicneuralnetwork} and fixed-precision quantization method PACT \citep{choi2018pact}. 
To ensure the baseline methods optimize for certified robustness, we added Gaussian noise to the inputs and used the accuracy metric on the Gaussian-noised inputs as described in Section \ref{sec:problem} during the policy search process, so that the quantization policy is learned based on Gaussian-noised images. For certifying the original smoothed classifier $g$ and quantized smoothed classifier $g^P$, we used RS on 500 images each with $10^6$ samples.

\noindent{\bf Experimental Setup and Hyperparameters.} For the ResNets experiments, we use a 48-core Intel Xeon Silver 4214R CPU with two Nvidia RTX A5000 GPUs. For the MobileNetV2 experiments, we use an AMD EPYC 7763 CPU with four Nvidia A100 GPUs. \Tool{} is implemented in Python and uses PyTorch.
We use SGD with a momentum of 0.9 and a weight decay of $10^{-4}$ for model training and fine-tuning following \citet{DBLP:conf/icml/CohenRK19}. 
During the policy search, we fine-tune the CIFAR-10 models for one epoch with a learning rate of 0.01, and the ImageNet models on a 60,000-sample subset with a learning rate of 0.001.
For fine-tuning in evaluation, for CIFAR-10, we set an initial learning rate of 0.01 and scaled it by 0.1 at epoch 5. 
For ImageNet, we set an initial learning rate of $10^{-3}$ and 
used the ReduceLROnPlateau learning rate scheduler. Fine-tuning is limited to 10 epochs, and the batch sizes are 256 for CIFAR-10 and 128 for ImageNet. 
A detailed description of our fine-tuning epoch choice is described in Section~\ref{sec:abl}.
For the optimization of the DDPG agent, following \citet{wang2019haq}, we use ADAM \citep{kingma2017adam} with $\beta_1 = 0.9$ and $\beta_2 = 0.999$. The learning rate is  $10^{-4}$ for the actor and $10^{-3}$ for the critic. During exploration, truncated normal noise with an initial standard deviation of 0.5, decaying at 0.99 per episode, is \mbox{applied to the actions.}
To account for statistical variation, we repeated each experiment in CIFAR-10 three times using different random seeds. We report the mean and 1-sigma error bars computed as sample standard deviation using \texttt{np.std(..., ddof=1)}.

\section{Experimental Results}

We present our main evaluation results: (1) the robustness and clean accuracy on the datasets;%
(2) the runtime of \Tool{}'s search algorithm; and (3) selected ablation studies.

\subsection{Robustness and Accuracy Evaluation on CIFAR-10}

On CIFAR-10, we used ResNet-20 as the base classifier with $\sigma = {0.25, 0.5, 1.0}$ and various BitOPs constraints. Figure \ref{fig:1} presents ACR vs. BitOps Pareto frontiers for the DNNs found by \Tool{} and the baseline tools.

\Tool{} achieved the best ACR for all experiments.
Figure \ref{fig:1a} presents results on $\sigma = 0.25$. \Tool{} consistently outperforms all baseline methods across different BitOPs constraints.
Figure \ref{fig:1b} presents results on $\sigma = 0.50$. Notably, \Tool{}'s 4-bit and 5-bit equivalent models outperformed the floating-point original model, which was not achieved by any other method.
Figure \ref{fig:1c} presents results on $\sigma = 1.00$. On 3-bit equivalent BitOPs constraint, \Tool{} had a 0.97\% ACR drop while the best baseline had a 3.43\% ACR drop. \Tool{} also outperformed the floating-point \mbox{model with 1.5\% operations.}

Figure \ref{fig:1d} shows the average clean accuracy drop achieved by the methods. ARQ's drop is smaller (hence, clean accuracy is higher) than the other baselines. Except for the 4-bit equivalent BitOPs, \Tool{} outperformed all baselines.

\begin{figure*}[h!]
  \centering
  \begin{subfigure}[b]{0.45\textwidth}
    \centering
    \includegraphics[width=.85\textwidth]{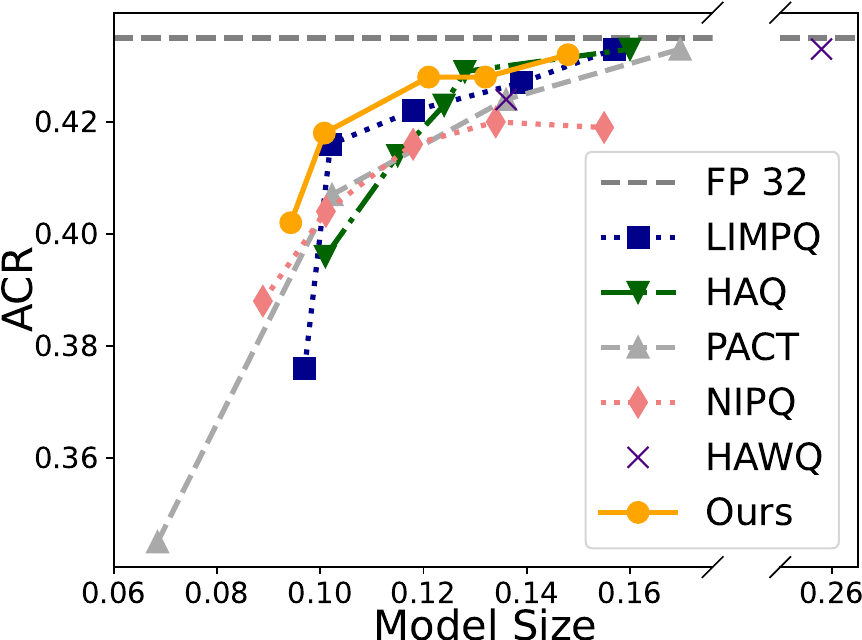}
    \caption{$\sigma = 0.25$}
    \label{fig:2a}
  \end{subfigure}
  \hfill
  \begin{subfigure}[b]{0.45\textwidth}
    \centering
    \includegraphics[width=.85\textwidth]{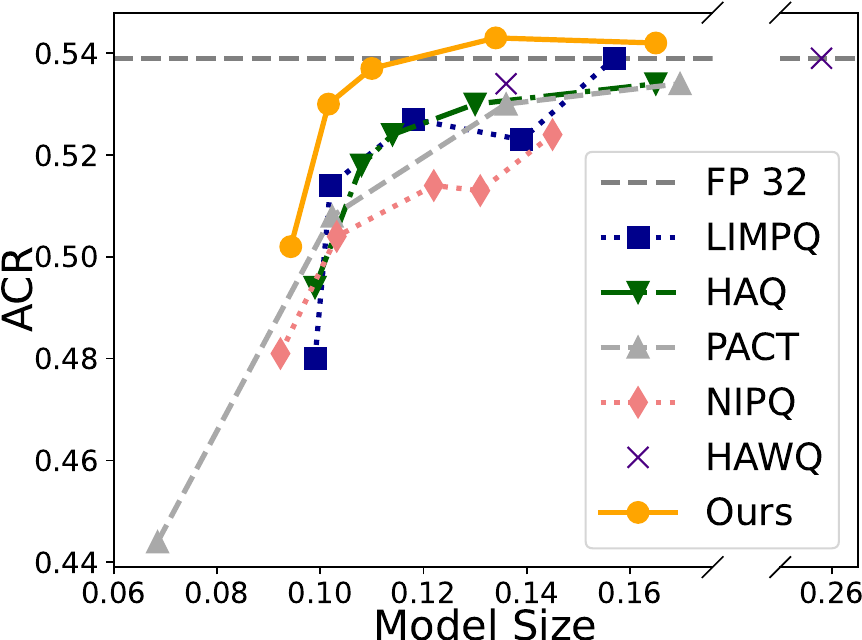}
    \caption{$\sigma = 0.5$}
    \label{fig:2b}
  \end{subfigure}
  
  \vspace{0.5em}
  
  \begin{subfigure}[b]{0.45\textwidth}
    \centering
    \includegraphics[width=.85\textwidth]{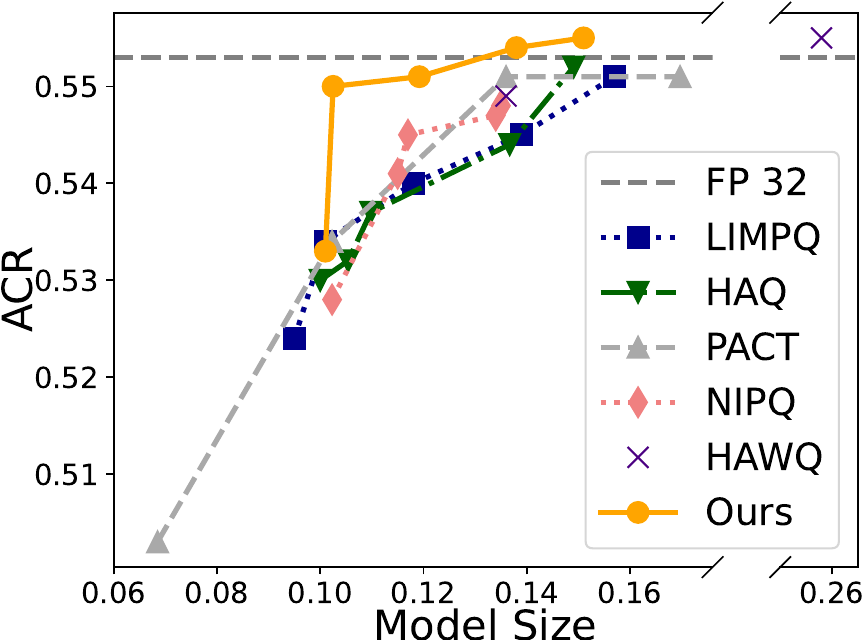}
    \caption{$\sigma = 1.0$}
    \label{fig:2c}
  \end{subfigure}
  \hfill
  \begin{subfigure}[b]{0.45\textwidth}
    \centering
    \includegraphics[width=.85\textwidth]{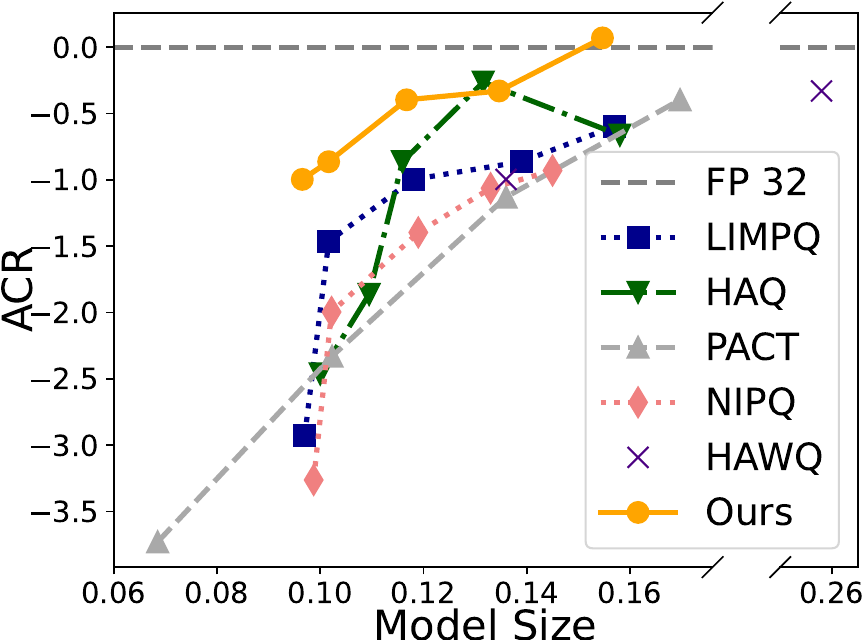}
    \caption{Avg. $\Delta$ Clean Acc}
    \label{fig:2d}
  \end{subfigure}
   \vspace{-.05in}
  \caption{Experiments on CIFAR-10. The x-axis shows the model size of $f_P$. The y-axis shows the ACR for the first three subfigures, and the average difference in clean accuracy between the methods and the original FP32 network across different $\sigma$ settings for Figure \ref{fig:2d}.}
  \vspace{-.1in}
  \label{fig:2}
\end{figure*}

While our primary experiments focused on using BitOPs as the constraint, here we present supplementary results with ACR and average difference in clean accuracy as the y-axis and model size as the x-axis. This provides additional insights into model size measurements, even though model size was not used as a constraint in our main experiments. Figure \ref{fig:2} compares \Tool{} with baseline methods.

\renewcommand{\arraystretch}{1.35}
\begin{table*}[ht]
  \caption{Experiments on ImageNet. The ACR denotes the average certified radius, Acc denotes the clean accuracy of the smoothed classifiers, BOPs  denotes the number of bit operations of the models (in G), {and Size is the model size (in MB).}}
  \label{tablei}
  \centering
  \begin{tabular}{l@{\hskip 0.08in}l@{\hskip 0.08in}c@{\hskip 0.08in}c@{\hskip 0.08in}c@{\hskip 0.08in}c@{\hskip 0.08in}c@{\hskip 0.08in}c@{\hskip 0.08in}c@{\hskip 0.08in}c@{\hskip 0.08in}c@{\hskip 0.08in}c@{\hskip 0.08in}c@{\hskip 0.08in}c}
    \toprule
    \multirow{2}{*}{ } & \multirow{2}{*}{Method} & \multicolumn{4}{c}{$\sigma = 0.25$} & \multicolumn{4}{c}{$\sigma = 0.50$} & \multicolumn{4}{c}{$\sigma = 1.00$} \\
    \cmidrule(lr){3-14}
    & & ACR & Acc & BOPs & {Size} & ACR & Acc & BOPs & {Size} & ACR & Acc & BOPs & {Size} \\
    \midrule
    \multirow{7}{*}{\rotatebox{90}{ResNet-50}} 
    & FP 32        & 0.488 & 69.4 & 4244.3 & {97.29} & 0.743 & 62.4 & 4244.3 & {97.29} & 0.914 & 45.4 & 4244.3 & {97.29} \\
    \cdashline{2-14}
    & \textbf{\Tool{}}  & \textbf{0.472} & \textbf{70.8} & \textbf{63.32} & {13.99} & \textbf{0.724} & \textbf{61.2} & \textbf{63.50} & \textbf{13.06} & \textbf{0.916} & \textbf{46.0} & \textbf{63.50} & {12.70} \\
    \cdashline{2-14}
    & LIMPQ           & 0.458 & 68.6 & 63.55 & \textbf{13.08} & 0.700 & 58.2 & 63.55 & {13.46} & 0.871 & 44.4 & 63.55 & {13.25} \\
    & {NIPQ}     & {0.460} & {68.4} & {63.56} & {13.21} & {0.696} & {58.8} & {63.56} & {13.07} & {0.885} & {45.0} & {63.56} & {12.96} \\
    & HAQ             & 0.460 & 69.0 & 63.56 & {13.14} & 0.715 & 60.8 & 64.15 & {13.35} & 0.880 & 44.8 & 63.55 & \textbf{11.93} \\
    & {HAWQ}     & {0.464} & {69.2} & {63.56} & {13.14} & {0.716} & {60.8} & {63.56} & {13.14} & {0.891} & {44.8} & {63.56} & {13.14} \\
    & PACT            & 0.460 & 69.0 & 63.56 & {13.14} & 0.715 & 61.0 & 63.56 & {13.14} & 0.884 & 45.4 & 63.56 & {13.14} \\
    \midrule
    \multirow{7}{*}{\rotatebox{90}{MobileNet-V2}} 
    & FP 32        & 0.457 & 67.0 & 308.24 & {13.24} & 0.668 & 57.0 & 308.24 & {13.24} & 0.846 & 44.4 & 308.24 & {13.24} \\
    \cdashline{2-14}
    & \textbf{\Tool{}}  & \textbf{0.385} & \textbf{62.0} & \textbf{4.60} & {2.23} & \textbf{0.576} & \textbf{54.0} & \textbf{4.60} & \textbf{2.13} & \textbf{0.774} & \textbf{41.6} & \textbf{4.60} & {2.27} \\
    \cdashline{2-14}
    & LIMPQ           & 0.347 & 58.4 & 4.62 & {2.16} & 0.540 & 50.6 & 4.62 & {2.24} & 0.703 & 40.2 & 4.62 & {2.16} \\
    & {NIPQ}     & {0.335} & {56.2} & {4.62} & \textbf{2.09} & {0.542} & {50.2} & {4.62} & {2.29} & {0.694} & {39.2} & {4.62} & \textbf{2.11 }\\
    & HAQ             & 0.341 & 56.0 & 4.60 & {2.27} & 0.573 & 53.4 & 4.60 & {2.24} & 0.683 & 39.8 & 4.61 & {2.17} \\
    & {HAWQ}     & {0.358} & {58.0} & {4.62} & {2.27} & {0.554} & {51.6} & {4.62} & {2.27} & {0.712} & {40.0} & {4.62} & {2.27} \\
    & PACT            & 0.376 & 60.2 & 4.62 & {2.27} & 0.564 & 52.4 & 4.62 & {2.27} & 0.757 & 40.2 & 4.62 & {2.27} \\
    \bottomrule
  \end{tabular}
\end{table*}

\subsection{Robustness and Accuracy Evaluation on ImageNet}

Table \ref{tablei} shows the results on ImageNet. We selected a $1.5\%$ BitOPs constraint as in CIFAR-10 \mbox{experiments}, which demonstrated that \Tool{} can achieve the accuracy and robustness of floating-point models.
\Tool{} outperformed all other quantization methods in all settings we consider. These results show that our approach with the ACR objective is able to improve \emph{both} the clean accuracy and robustness. 
For ResNet-50, the ACR of the ARQ-generated network is comparable to the ACR of the original floating-point model. The clean accuracy for $\sigma=0.25$ and $\sigma=1$ is even slightly higher than the accuracy of the original network.
As a result of the limited fine-tuning, for MobileNetV2, the clean accuracy and ACR are reduced compared to the floating-point model, but both are significantly higher than the alternative quantization methods. 
Since low $\sigma$ certifies small radii with high accuracy but not large radii, while high $\sigma$ certifies larger radii but with lower accuracy for smaller radii, the clean accuracy drops as $\sigma$ increases. This observation is consistent with that of RS \citep{DBLP:conf/icml/CohenRK19}.

The results further show that the other mixed-precision quantization methods could hardly outperform the ACR of fixed-precision PACT \citep{choi2018pact} on ImageNet, while \Tool{} significantly outperformed PACT on both networks and even the original floating-point model at \mbox{$\sigma = 1.00$ on ResNet-50.}

\subsection{Generalization to Transformer Architectures}
\renewcommand{\arraystretch}{1.25}
\begin{table*}[t]
  \caption{Experiments on CIFAR-10 with ViT-tiny and $\sigma=0.5$. We report the compute cost in BOPs (in G), the average certified radius (ACR), and the certified accuracy (\%) at different radius thresholds $r$ for the smoothed classifier. \Tool{} matches the BOPs budget of a 4-bit PACT baseline and is compared against the original FP32 model, demonstrating that \Tool{} extends beyond CNNs to transformer-based architectures.}
  \label{table_vit}
  \centering\protect\small
  \begin{tabular}{l@{\hskip 0.2in}c@{\hskip 0.2in}c@{\hskip 0.2in}c@{\hskip 0.2in}c@{\hskip 0.2in}c@{\hskip 0.2in}c@{\hskip 0.2in}c@{\hskip 0.2in}c@{\hskip 0.2in}c@{\hskip 0.2in}c@{\hskip 0.2in}c@{\hskip 0.2in}}
    \toprule
    \multirow{2}{*}{Method} & \multirow{2}{*}{BOPs} & \multirow{2}{*}{ACR} & \multicolumn{8}{c}{Radius r} \\
    \cmidrule(lr){4-11}
    & & & 0.0 & 0.25 & 0.50 & 0.75 & 1.00 & 1.25 & 1.50 & 1.75 \\
    \midrule
    FP 32 & 373.87 & 0.406 & 53.8 & 42.6 & 32.2 & 24.6 & 17.2 & 10.0 & 6.4 & 2.8 \\
    \hdashline
    \textbf{\Tool{}} & \textbf{25.45} & 0.388 & 51.4 & 40.2 & 32.8 & 23.0 & 17.0 & 9.6 & 5.0 & 2.8 \\
    \hdashline
    4-bit PACT & 25.45 & 0.369 & 49.4 & 39.4 & 29.4 & 21.4 & 15.0 & 8.6 & 4.6 & 2.6 \\
    \bottomrule
  \end{tabular}
\end{table*}

To evaluate the generalization capability of ARQ to modern transformer-based architectures, we conducted experiments on ViT-tiny~\citep{dosovitskiy2021imageworth16x16words} on CIFAR-10 with $\sigma=0.5$. Since our evaluation of multiple baseline methods across various DNN architectures showed consistent behavioral patterns, we selected PACT as the representative baseline for ViT experiments. Table~\ref{table_vit} shows that ARQ achieves ACR of 0.388, outperforming 4-bit PACT (0.369) under the same BitOPs constraint (25.45), while using 14.7x fewer operations compared to FP32. This demonstrates that ARQ's approach generalizes effectively beyond CNNs to transformer-based models. The IRS acceleration technique, being model-agnostic, applies equally to transformer blocks as it does to convolutional layers.

\subsection{Inference Latency Analysis}

To evaluate the practical deployment benefits of ARQ, we estimate the inference latency on the BISMO cloud accelerator~\citep{umuroglu2018bismoscalablebitserialmatrix} using the layer-wise latency lookup-table methodology from HAQ~\citep{wang2019haq}. Table~\ref{table_latency} shows the latency estimates for MobileNetV2 with batch size 16. ARQ achieves substantially higher ACR (0.774) compared to HAQ (0.683) at similar latency (17.12ms vs. 17.19ms) and BitOP budget. Compared to higher-precision settings like 5-bit PACT, ARQ achieves comparable robustness (0.774 vs. 0.776) with significantly lower latency (17.12ms vs. 21.55ms), demonstrating the efficiency gains from our mixed-precision quantization approach.

\begin{table}[h]
\centering
\caption{Inference latency estimates on ImageNet with MobileNetV2 using the BISMO cloud accelerator (batch size 16). We report the ACR, estimated inference latency per batch (ms), and BOPs (in G). \Tool{} achieves higher ACR than comparable-budget baselines with similar estimated latency.}
\label{table_latency}
\setlength{\tabcolsep}{12pt}
\begin{tabular}{lccc}
\toprule
Method & ACR & Latency (ms) & BOPs \\
\midrule
\textbf{ARQ} & 0.774 & 17.12 & 4.60 \\
\hdashline
HAQ & 0.683 & 17.19 & 4.61 \\
4-bit PACT & 0.757 & 16.55 & 4.62 \\
5-bit PACT & 0.776 & 21.55 & 7.21 \\
\bottomrule
\end{tabular}
\end{table}

\subsection{Comparison Experiments with Robustness-aware Quantization Methods}

Table \ref{table_ICR} compares the accuracy loss of \Tool{} and ICR \cite{lin2021integerarithmeticonlycertifiedrobustnessquantized} relative to FP32 models. Our experiments show that the 8-bit equivalent model of \Tool{} outperforms ICR's 8-bit model in both clean and robust accuracy.

\renewcommand{\arraystretch}{1.1}
\begin{table}[ht]
  \caption{Comparison with ICR on CIFAR-10 with ResNet-20 and $\sigma = 0.5$. We report the change in certified accuracy (\%) at each radius threshold $r$ relative to the FP32 baseline.}
  \label{table_ICR}
  \centering\protect
  \begin{tabular}{l@{\hskip 0.1in}c@{\hskip 0.1in}c@{\hskip 0.1in}c@{\hskip 0.1in}c@{\hskip 0.1in}c@{\hskip 0.1in}c@{\hskip 0.1in}c@{\hskip 0.1in}c@{\hskip 0.1in}c@{\hskip 0.1in}}
    \toprule
    \multirow{2}{*}{Method} & \multirow{2}{*}{BOPs} & \multicolumn{7}{c}{ $\Delta$ Accuracy on Radius r} \\
    \cmidrule(lr){3-9}
    & & \quad{}0.0 & \quad{}0.25 & \quad{}0.50 & \quad{}0.75 & \quad{}1.00 & \quad{}1.25 & \quad{}1.50\\
    \midrule
   \textbf{\Tool{}}  & \textbf{2.588} & \textbf{0.2} & \textbf{0.4} & \textbf{0.2} & \textbf{0.4} & \textbf{0.6} & \textbf{1.0} & \textbf{0.6} \\
    ICR           & 2.596 & -2.0 & -2.0 & -4.0 & -3.0 & 0 & 0 & -1.0 \\
    \bottomrule
  \end{tabular}
\end{table}

While empirical robustness methods are effective against known attacks, it has been shown that stronger adversaries often render them ineffective~\citep{tramèr2020ensembleadversarialtrainingattacks, athalye2018robustnesscvpr2018whitebox, carlini2016defensivedistillationrobustadversarial}. This shows the limitations of empirical robustness: it lacks formal guarantees and remains vulnerable to new attacks. In contrast, certified robustness provides guarantees against all perturbations within a given norm. This ensures worst-case robustness rather than best-effort defenses.
And certified robustness covers a wider range of adversarial attacks, including those not explicitly tested in empirical evaluations.

{\subsection{Comparing quantization aimed for certified and empirical robustness.} Typically, methods focusing on empirical robustness do not achieve high certified accuracy. To illustrate, we performed additional experiments using an 8-bit quantized ATMC~\citep{gui2019model} ResNet-20 model. Table~\ref{table_empirical} shows that methods optimized for empirical robustness, such as ATMC, offer minimal gains in certified robustness, expressed as ACR, compared to our ARQ method, which yields ACR that is 3.5--13.9$\times$ larger. In fact, these empirically robust methods have almost no ability to correctly classify inputs perturbed with Gaussian noise, which is crucial for robustness certification, and perform similarly to a randomly initialized model.}

\renewcommand{\arraystretch}{1.1}
\begin{table}[H]
  \caption{Certified robustness vs.\ empirical robustness under the randomized smoothing evaluation. We compare \Tool{} against an 8-bit ATMC baseline on CIFAR-10 with ResNet-20 and report ACR across different noise levels $\sigma$.}
  \vspace{-.1in}
  \label{table_empirical}
  \centering\small
  \setlength{\tabcolsep}{12pt}
  \begin{tabular}{lccc}
    \toprule
    \multirow{2}{*}{Method} & \multicolumn{3}{c}{ACR} \\
    \cmidrule(lr){2-4}
    & $\sigma = 0.25$ & $\sigma = 0.50$ & $\sigma = 1.00$ \\
    \midrule
    \textbf{\Tool{}}  & \textbf{0.432} & \textbf{0.543} & \textbf{0.555} \\
    ATMC~\citep{gui2019model} & 0.031 & 0.072 & 0.159 \\
    \bottomrule
  \end{tabular}
\end{table}

\subsection{Execution Time of \Tool{} Search and Other Tools}

Table \ref{tablet} presents the time consumption for different methods. 
We selected $\sigma=0.5$ as it is the median value among the tested values in the experiments.
The total time includes the policy search, fine-tuning, and evaluation.  
The ``Eval'' time column is the time for fine-tuning and evaluation of the quantized smoothed classifiers.
Although \Tool{} consumes more time than other methods, it is an offline, one-time cost: the policy search is run once to produce a deployable mixed-precision model, and no search is performed during inference. When the model is deployed for extended inference, this cost can be amortized across many inference runs.

\renewcommand{\arraystretch}{1.1}
\begin{table}[ht]
  \caption{Execution time for \Tool{} and baseline quantization search methods, measured in hours. We report the policy search time and evaluation time for each method.}
  \label{tablet}
  \centering
  \setlength{\tabcolsep}{12pt}
  \begin{tabular}{ccccccc}
    \toprule
    \multirow{2}{*}{Benchmark} & \multicolumn{4}{c}{Policy Search} & \multirow{2}{*}{Eval} \\
    \cmidrule(lr){2-5}
    & \Tool{} & HAQ & LIMPQ & NIPQ  \\
    \midrule
    ResNet-50  & 71.89  &  42.65  &  23.74 & 121.39 & 32.86   \\
    MobileNetV2 &   67.83   &   29.37   &   14.35 & 78.31 &  17.61       \\
    ResNet-20  &   2.83   &   2.31   &   0.09 & 0.717 &   0.47       \\
    \bottomrule
  \end{tabular}
\end{table}

Table \ref{table_time} presents the results of \Tool{} under different time constraints for policy search. This information serves as a reference for users aiming to balance time consumption in policy optimization with accuracy and robustness during inference. We include HAQ, the strongest baseline method with the highest ACR among all baselines, as a reference. HAQ achieves an ACR of 0.518 with a search time of 2.31 hours. Notably, although \Tool{}'s performance declines with reduced search time, it consistently outperforms this strongest baseline across all time budgets. Even with the most aggressive time constraint of 0.25 hours (approximately 10\% of HAQ's search time), \Tool{} achieves an ACR of 0.521, surpassing HAQ's result. This demonstrates \Tool{}'s efficiency in finding high-quality quantization policies even under severe time limitations.

\renewcommand{\arraystretch}{1.1}
\begin{table*}[ht]
  \caption{Effect of policy search time on \Tool{} for CIFAR-10 with ResNet-20 and $\sigma=0.5$. We report the resulting model cost, ACR and certified accuracy at different radius thresholds $r$.}
  \label{table_time}
  \centering\protect
  \begin{tabular}{l@{\hskip 0.2in}c@{\hskip 0.2in}c@{\hskip 0.2in}c@{\hskip 0.2in}c@{\hskip 0.2in}c@{\hskip 0.2in}c@{\hskip 0.2in}c@{\hskip 0.2in}c@{\hskip 0.2in}c@{\hskip 0.2in}c@{\hskip 0.2in}c@{\hskip 0.2in}}
    \toprule
    \multirow{2}{*}{Method} & \multirow{2}{*}{Time} & \multirow{2}{*}{BOPs} & \multirow{2}{*}{ACR} & \multicolumn{8}{c}{Radius r} \\
    \cmidrule(lr){5-12}
    & & & & 0.0 & 0.25 & 0.50 & 0.75 & 1.00 & 1.25 & 1.50 & 1.75 \\
    \midrule
    \multirow{3}{*}{\textbf{\Tool{}}} & 2.83 & 0.354 & 0.530 & 67.2 & 54.6 & 43.2 & 32.6 & 22.2 & 14.2 & 7.4 & 4.4 \\
      & 2.18 & 0.354 & 0.528 & 66.8 & 54.4 & 42.8 & 32.4 & 22.0 & 14.0 & 7.2 & 4.2 \\
      & 0.25 & 0.354 & 0.521 & 66.2 & 53.8 & 43.0 & 32.0 & 21.8 & 13.8 & 7.2 & 4.0 \\
    \hdashline
    HAQ & 2.31 & 0.363 & 0.518 & 66.4 & 53.4 & 43.0 & 32.8 & 21.6 & 13.0 & 7.4 & 4.0 \\
    LIMPQ & 0.09 & 0.361 & 0.514 & 65.0 & 52.0 & 41.8 & 31.4 & 20.8 & 13.2 & 7.4 & 4.0 \\
    \bottomrule
  \end{tabular}
\end{table*}

\subsection{Ablation Studies}\label{sec:abl}

\textbf{Choice of Reward Function.} We investigate the sensitivity of the policy search to the quality of the reward function. In \Tool{}, ACR is used as a reward for the RL agent. An intuitive question arises: What if we use the certified accuracy of the quantized smoothed classifier $g_P$ as the reward? As noted in Section \ref{sec:cnr}, certified accuracy is the probability that $g_P$ correctly predicts samples $x$ with certified radius $R(x)$ that exceeds the given threshold~$r$.
Table~\ref{table ablation} presents the ablation study results on 3-bit equivalent quantized ResNet-20 models. "Val" and "Acc" denote using the validation accuracy of $f_P$ and the clean accuracy of $g_P$. "ACR+Acc" combines ACR with the accuracy of $g_P$ as the reward function. $\text{Acc}_{0.5}$ represents the certified accuracy of $g_P$ at $r = 0.5$. We observe that "Acc" achieves better-certified accuracy at the targeted $r$ but underperforms at other radii and in ACR.

\renewcommand{\arraystretch}{1.1}
\begin{table*}[h]
  \caption{Reward function ablation on CIFAR-10 with ResNet-20 and $\sigma = 0.5$. Each row corresponds to a different reward definition used during policy search, and we report the resulting ACR and certified accuracy at different radius thresholds $r$.}
  \label{table ablation}
  \centering\protect
  \begin{tabular}{l@{\hskip 0.2in}c@{\hskip 0.2in}c@{\hskip 0.2in}c@{\hskip 0.2in}c@{\hskip 0.2in}c@{\hskip 0.2in}c@{\hskip 0.2in}c@{\hskip 0.2in}c@{\hskip 0.2in}c@{\hskip 0.2in}c@{\hskip 0.2in}}
    \toprule
    \multirow{2}{*}{Method} & \multirow{2}{*}{BOPs} & \multirow{2}{*}{ACR} & \multicolumn{8}{c}{Radius r} \\
    \cmidrule(lr){4-11}
    & & & 0.0 & 0.25 & 0.50 & 0.75 & 1.00 & 1.25 & 1.50 & 1.75 \\
    \midrule
    FP 32        & 42.04 & 0.539 & 68.2 & 56.0 & 44.6 & 33.8 & 21.8 & 14.4 & 7.2 & 3.8 \\
        \hdashline
   \textbf{\Tool{}}  & \textbf{0.354} & \textbf{0.530} & \textbf{67.2} & 54.6 & 43.2 & 32.6 & \textbf{22.2} & \textbf{14.2} & \textbf{7.4} & \textbf{4.4} \\
    \hdashline
    Val           & 0.363 & 0.518 & 66.4 & 53.4 & 43.0 & \textbf{32.8} & 21.6 & 13.0 & \textbf{7.4} & 4.0 \\
    {Acc} & {0.360} & {0.512} & {67.0} & {53.2} & {42.6} & {31.4} & {20.4} & {13.2} & {6.8} & {3.6} \\
    $\text{Acc}_{0.5}$ & 0.362 & 0.525 & 65.8 & \textbf{55.6} & \textbf{43.8} & 31.8 & 21.6 & 13.0 & \textbf{7.4} & 4.0 \\
    {ACR+Acc} & {0.362} & {0.526} & {\textbf{67.2}} & {54.2} & {\textbf{43.8}} & {31.6} & {21.2} & {13.6} & {7.2} & {4.2} \\
    \bottomrule
  \end{tabular}
\end{table*}

\textbf{Effect of Incremental RS.}
In our experiments, IRS achieves a 1.32x speedup over RS, aligning with~\cite{IRS}. Table \ref{table_IRS} shows the detailed results of using RS instead of IRS to obtain $ACR_p$ and $Reward_t$.  Moreover, using IRS does not reduce the quality of the quantization policy (and in many cases improves it), as ARQ's algorithm design makes subsequent candidate quantized networks similar enough to benefit from incremental robustness proving. 

\renewcommand{\arraystretch}{1.25}
\begin{table*}[ht]
  \caption{Effect of incremental randomized smoothing, IRS, versus rerunning RS on CIFAR-10. The results are reported across different noise levels $\sigma$.}
  \label{table_IRS}
  \centering
  \begin{tabular}{l@{\hskip 0.2in}c@{\hskip 0.2in}c@{\hskip 0.2in}c@{\hskip 0.2in}c@{\hskip 0.2in}c@{\hskip 0.2in}c@{\hskip 0.2in}c@{\hskip 0.2in}c@{\hskip 0.2in}c@{\hskip 0.2in}}
    \toprule
    \multirow{2}{*}{Method} & \multicolumn{3}{c}{$\sigma = 0.25$} & \multicolumn{3}{c}{$\sigma = 0.50$} &\multicolumn{3}{c}{$\sigma = 1.00$} \\
    \cmidrule(lr){2-10}
    & ACR  & Acc & BOPs & ACR & Acc & BOPs & ACR & Acc & BOPs \\
    \midrule
    FP 32        & 0.435 & 76.4 & 42.04 & 0.539 & 68.2 & 42.04 & 0.553 & 50.8 & 42.04 \\
        \hdashline
   \textbf{\Tool{}}(-IRS)  &  \textbf{0.418} & \textbf{76.0} & \textbf{0.362} & \textbf{0.530} & \textbf{67.2} & \textbf{0.354} & \textbf{0.550} & \textbf{49.6} & \textbf{0.354} \\
    \hdashline
    \Tool{}-RS           & 0.414 & 75.8 & \textbf{0.362} & 0.526 & \textbf{67.2} & 0.359 & 0.537 & 49.2 & 0.359 \\
    \bottomrule
  \end{tabular}
\end{table*}

\begin{figure}[h]
  \centering
    \includegraphics[width=.75\textwidth]{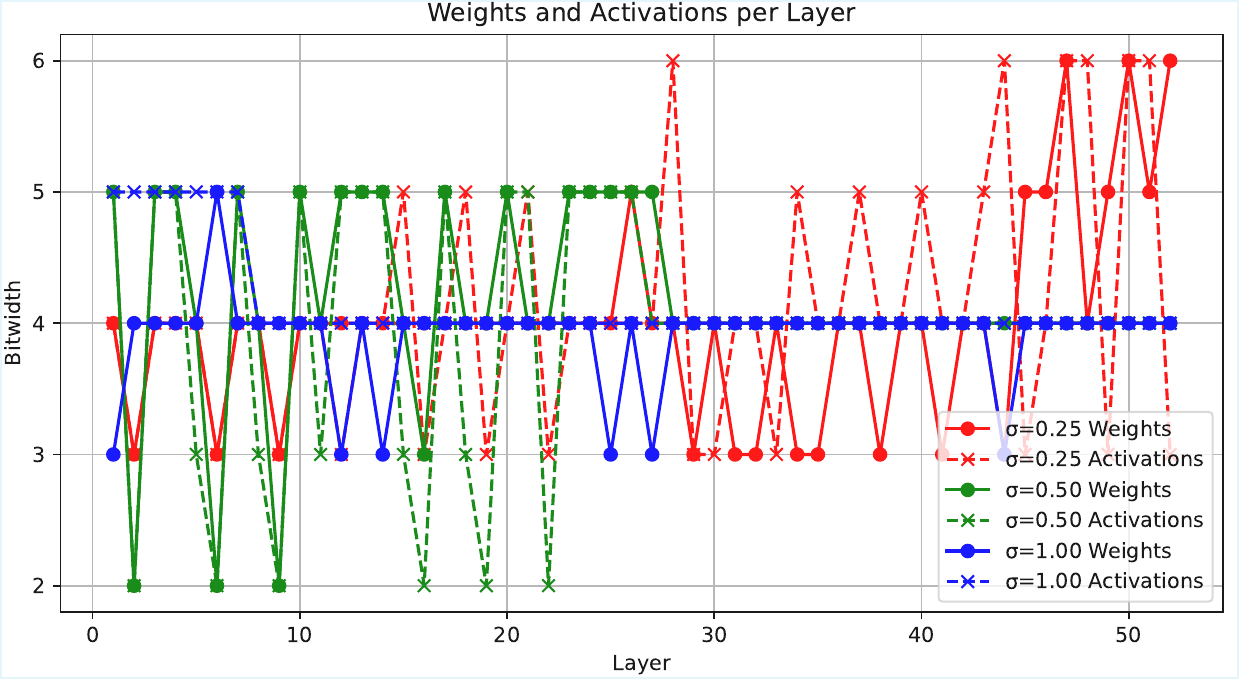}
    \vspace{-.1in}
  \caption{Quantization policy among different $\sigma$s for ResNet-50 on ImageNet. The x-axis represents the layer index, and the y-axis represents the bit-width selection in the quantization policy for each specific layer. The \(\bullet\) symbol represents the bit-widths for weights, and the \(\times\) symbol represents the bit-widths for activations. The figure highlights how the layer-wise precision allocation changes as $\sigma$ varies.}
  \label{fig:b} %
\end{figure}

\textbf{Quantization Bit-width for Different Layers.}
ARQ effectively learns mixed-precision quantization policies with different $\sigma$ values leading to distinct policies. 
Figure \ref{fig:b} illustrates the significance of generating different quantization policies for different values of $\sigma$. The quantization policy for $\sigma=0.25$ is more aggressive in the first half of the layers and more conservative in the later layers. This pattern is different from that of $\sigma=0.5$ and $\sigma=1.0$, where the policy tends to make more adjustments in the first half and remains stable in the latter half.

\textbf{Effect of the Number of Epochs in Fine-tuning.}
Extending fine-tuning from 10 to 90 epochs maintains the observed trends in ACR and accuracy.
Due to the limitation of computational resources, we only conducted fine-tuning for 10 epochs in our experiments. Table \ref{table_finetune} shows the results of further fine-tuning for a total of 90 epochs. The trends in ACR and clean accuracy followed the same pattern as observed during the initial 10 epochs, demonstrating the effectiveness of ARQ.

\renewcommand{\arraystretch}{1.25}
\begin{table}[t]
  \caption{Effect of fine-tuning length on CIFAR-10 with ResNet-20 and $\sigma=0.5$. We compare results after 10 and 90 fine-tuning epochs, using the same table format as Table~\ref{tablei}.}
  \label{table_finetune}
  \centering
  \small
  \setlength{\tabcolsep}{14pt}
  \begin{tabular}{lccccc}
    \toprule
    \multirow{2}{*}{Method} & \multirow{2}{*}{BOPs} & \multicolumn{2}{c}{$\text{Epochs} = 90$} & \multicolumn{2}{c}{$\text{Epochs} = 10$}  \\
    \cmidrule(lr){3-6}
    & & ACR  & Acc & ACR & Acc \\
    \midrule
   \textbf{\Tool{}}  &  \textbf{0.354} & \textbf{0.545} & \textbf{67.6} & \textbf{0.530} & \textbf{67.2} \\
   \hdashline
    LIMPQ           & 0.361 & 0.535 & 67.0 & 0.514 & 65.0 \\
    HAQ             & 0.365 & 0.534 & 66.8 & 0.518 & 66.4\\
    PACT            & 0.362 & 0.524 & 66.6 & 0.508 & 65.8\\
    \bottomrule
  \end{tabular}
\end{table}

\textbf{RL Hyperparameter Sensitivity.}
ARQ's sensitivity to the RL actor's learning rate shows resilience across different settings.
We evaluated the sensitivity of ARQ to the RL actor's learning rate. Table~\ref{table:actor_lr} shows how different learning rates affect the ACR and clean accuracy. The results show the resilience of ARQ to the choice of RL hyperparameters.
\renewcommand{\arraystretch}{1.25}
\begin{table*}[ht]
    \caption{Sensitivity to the actor learning rate in \Tool{} on CIFAR-10 with ResNet-20 and $\sigma=0.5$. We report certified accuracy at different radius thresholds $r$ for the models found under each learning rate.}
    \label{table:actor_lr}
    \centering\protect
    \small
    \begin{tabular}{c@{\hskip 0.2in}c@{\hskip 0.2in}c@{\hskip 0.2in}c@{\hskip 0.2in}c@{\hskip 0.2in}c@{\hskip 0.2in}c@{\hskip 0.2in}c@{\hskip 0.2in}c@{\hskip 0.2in}c@{\hskip 0.2in}c@{\hskip 0.2in}}
    \toprule
    \multirow{2}{*}{Actor LR} & \multirow{2}{*}{BOPs} & \multirow{2}{*}{ACR} & \multicolumn{8}{c}{Radius r} \\
    \cmidrule(lr){4-11}
    & & & 0.0 & 0.25 & 0.50 & 0.75 & 1.00 & 1.25 & 1.50 & 1.75 \\
    \midrule
    1e-3 & 0.355 & 0.524 & 66.8 & 54.0 & 42.6 & 32.2 & 21.8 & 13.8 & 7.2 & 4.0 \\
    \textbf{1e-4} & 0.354 & 0.530 & 67.2 & 54.6 & 43.2 & 32.6 & 22.2 & 14.2 & 7.4 & 4.4 \\
    1e-5 & 0.354 & 0.521 & 66.6 & 53.6 & 42.8 & 31.8 & 21.8 & 13.6 & 7.2 & 3.8 \\
    \bottomrule
    \end{tabular}
    \vspace{-.15in}
\end{table*}

\section{Threats to Validity}

\noindent\textbf{Internal.} \Tool{} employs DDPG as the reinforcement learning algorithm for policy search. While we use a specific RL algorithm, the core idea of integrating incremental randomized smoothing into the quantization search loop is general and can be applied with other RL or search algorithms. We use BitOps as a surrogate metric for performance, which may not represent savings on existing hardware, but this risk is mitigated by also presenting the model size showing high correlation (e.g., Figure~\ref{fig:2}) and by the fact that modern accelerators start to incorporate low-precision arithmetic~\cite{lowbit_fpga_1, lowbit_fpga_2,9689050,tahmasebi2025flexibitfullyflexibleprecision}. 

\noindent\textbf{External.} Our evaluation uses specific neural network architectures (ResNet-20, ResNet-50, MobileNetV2, and ViT-tiny) and datasets (CIFAR-10 and ImageNet). While we do not evaluate all possible networks, the selected models are widely used benchmarks in both quantization and robustness research. Similarly, we focus on representative quantization methods covering both mixed-precision (HAQ, LIMPQ, NIPQ, HAWQ-V3) and fixed-precision (PACT) approaches.

\noindent\textbf{Construct.} We use $\ell_2$-norm perturbations with Gaussian noise following the standard randomized smoothing setup. While real-world adversarial attacks may use other perturbation types (e.g., $\ell_\infty$), our approach can be extended to other noise distributions that certify different threat models~\citep{yang2020randomized}.

\noindent\textbf{Statistical.} Randomized smoothing is a statistical method, so there is a small probability (controlled by the confidence parameter $\alpha$) that the certified bounds may not hold. We use $\alpha=0.001$ following prior work~\citep{DBLP:conf/icml/CohenRK19, IRS}, corresponding to 99.9\% confidence.

\section{Related Work}

\noindent{\bf Mixed-Precision Quantization.}
To optimize the balance between the accuracy and efficiency of DNNs, many mixed-precision quantization methods have been presented, and quantization-based efficient inference designs have also been explored~\citep{yang2025neubridge}.
\citet{dong2019hawq, louizos2017bayesian, chen2021mixedprecision, tang2023mixedprecision} employ appropriate proxy metrics that indicate model sensitivity to quantization to generate quantization policies.
Some other researchers formulated quantization policy optimization as a search problem and addressed it using a Markov Decision Process through reinforcement learning \citep{wang2019haq, lou2020autoq, elthakeb2020releq} and a differentiable search process employed Neural Architecture Search algorithms \citep{guo2020single, wu2018mixed}. Related automation efforts in systems have also targeted efficient quantization kernels on GPUs~\citep{zhao2026nautilus}. 
As Table \ref{tablei} shows, \Tool{} outperformed HAQ \citep{wang2019haq} and LIMPQ \citep{tang2023mixedprecision}, which are state-of-the-art mixed-precision quantization methods.

\noindent{\bf Certified and Empirical Robustness Techniques.}
\textit{Certified robustness techniques}
\citep{raghunathan2018certified,li2023sok, DBLP:journals/pacmpl/UgareSM22,
ugare2023incremental} provide sound, formal guarantees that for all perturbed inputs in a set,
the DNN prediction (label in classification tasks) remains unchanged,
but these techniques incur a high computational cost.
There are two categories of certified robustness techniques: deterministic and statistical,
whereas deterministic techniques
\citep{singh2019boosting,
  lechner2022quantizationawareintervalboundpropagation, 10.1145/3597926.3598127,
10.1145/3540250.3558924, zhang22babattack,laurel2023synthesizing,yang2022provable}
are precise but more expensive.
Statistical methods~\citep{DBLP:conf/icml/CohenRK19,
li2021tsstransformationspecificsmoothingrobustness, yang2022certifiedrobustnessensemblemodels}
are more scalable and often evaluated on networks that are too \mbox{large for deterministic methods.}
\textit{Empirical robustness
techniques}~\citep{prakash2018deflectingadversarialattackspixel,adv-training-cvpr2024,safer-iccv2025}
use concrete adversarial inputs to evaluate and improve the robustness of the model.
They are effective against specific attack inputs but do not provide guarantees,
and are only as strong as the evaluation (threat model, attack strength, dataset coverage, etc.).
The conceptual difference between empirical and certified robustness is discussed in Section~\ref{sec:cnr}.
As empirical and certified robustness techniques differ in definition of robustness,
applicable DNNs, and evaluation methods, \mbox{these techniques are rarely comparable.
}

\noindent{\bf Robustness of Quantized Models.}
{Recent studies have systematically characterized how quantization affects model behavior and reliability. \citet{hu2022characterizingunderstandingbehaviorquantized} conducted an empirical study revealing that quantization can introduce behavioral discrepancies that may degrade reliability in deployment. DiffChaser~\citep{ijcai2019p800} detects disagreements between a reference model and its variants, 
while DiverGet~\citep{Yahmed_2022} provides a search-based testing approach specifically for quantization assessment. These works motivate the need for robustness-aware quantization methods like ARQ.}

\renewcommand{\arraystretch}{1.15}
\begin{table}[t]
\small
  \caption{Comparison of various robustness-aware model reduction methods. \textbf{\textit{Model Compression Method:}}  P -- Pruning; SPQ -- Single-Precision Quantization; MPQ -- Mixed-Precision Quantization. \textbf{\textit{Stage:}} T~--~Training; PT -- Post-training (tuning). \textit{\textbf{Model Size}}: Largest supported data-set in the paper's evaluation.}
  \label{table_RW}
  \centering\vspace{-.1in}
  \begin{tabular}{l@{\hskip 10pt}c@{\hskip 10pt}c@{\hskip 10pt}c@{\hskip 10pt}c@{\hskip 10pt}c@{\hskip 10pt}c@{\hskip 10pt}c}

    \toprule
    \multirow{2}{*}{Approach} & \multicolumn{7}{c}{Properties} \\
    \cmidrule(lr){2-8}
    & \makecell{Randomized \\ Smoothing} & \makecell{Deterministic \\ Certification} & \makecell{Empirical \\ Robustness} & \makecell{Compression \\ Method} & \makecell{Compression \\Stage} & \makecell{Max Evaluated  \\ Model Size} \\
    \midrule
    ARQ & \checkmark &  &  & MPQ & PT & ImageNet \\\hdashline
    \href{https://arxiv.org/abs/1902.03538}{ATMC}  &  &  & $\checkmark$  & SPQ & T & CIFAR-100 \\
    \href{https://arxiv.org/abs/1904.08444}{DQ} &  &  & $\checkmark$  & SPQ & T & CIFAR-10 \\
    \href{https://arxiv.org/abs/2002.07520}{GRQR} &  &  & $\checkmark$  & SPQ & PT & ImageNet \\
    \href{https://arxiv.org/abs/2108.09413}{ICR} & $\checkmark$ &  &  & SPQ & T & Caltech-101 \\
    \href{https://arxiv.org/abs/2211.16187}{QIBP} &  & $\checkmark$ &  & SPQ & T & CIFAR-10 \\
    \href{https://arxiv.org/abs/1903.12561}{ARMC} &  &  & $\checkmark$  & P & T & CIFAR-10 \\
    \href{https://arxiv.org/abs/2004.11233}{QUANOS} &  &  & $\checkmark$  & MPQ & PT & CIFAR-100 \\
    \href{https://arxiv.org/abs/2105.06512}{S-Shield} &  &  & $\checkmark$  & SPQ & PT & CIFAR-10 \\
    \href{https://arxiv.org/abs/2002.10509}{HYDRA} & $\checkmark^*$ & $\checkmark^*$ & $\checkmark$  & P & PT & CIFAR-10 \\ %
    \href{https://arxiv.org/abs/1906.06110}{TCR} &  &  & $\checkmark$  & P & PT & CIFAR-10 \\
    \href{https://arxiv.org/abs/2011.03083}{DNR} &  &  & $\checkmark$  & P & T & Tiny-ImageNet \\
    \href{https://link.springer.com/chapter/10.1007/978-3-030-45237-7_5}{HMBDT}  &  & $\checkmark$ &  & SPQ & PT & MNIST \\
    \href{https://doi.org/10.1609/aaai.v34i04.6105}{TCMR} &  & $\checkmark$ &  & SPQ & PT & CIFAR-10 \\
    \bottomrule
  \end{tabular}
\vspace{-.1in}
\end{table}

Several existing works address the challenge of training DNNs for both efficiency and robustness,
targeting safety-critical and resource-limited applications.
ATMC~\citep{gui2019model} unified various existing compression techniques.
DQ~\citep{lin2019defensive} and
GRQR~\citep{alizadeh2020gradientell1regularizationquantization} presented how controlling
the magnitude of adversarial gradients can be used to construct a defensive quantization method.
We summarize these methods and compare them to \Tool{} in Table~\ref{table_RW}.
As illustrated in Table~\ref{table_RW}, \Tool{} stands out by being the only approach that combines mixed-precision quantization with post-training optimization for certified robustness on large-scale datasets like ImageNet. This distinguishes our work from other approaches that either focus only on pruning (which removes over 90\% of the weights) and often target smaller datasets. Although ICR~\citep{lin2021integerarithmeticonlycertifiedrobustnessquantized} proposed quantization methods with RS during training, \Tool{} focuses on post-training optimization and scales to ImageNet (compared to ICR's much smaller Caltech-101). Section~\ref{sec:abl} provides a detailed comparison between ICR and \Tool{}. HYDRA~\citep{sehwag2020hydrapruningadversariallyrobust} analyzes pruning with RS but only scales to CIFAR-10 for RS and is hard to transfer to quantization methods.

{\noindent{\bf  Testing and Analysis for Quality Assurance.}
Previous research in approximate computing made the observation that in many application domains, one can optimize programs in a multi-objective tradeoff space, with (soft) accuracy and (stricter) safety/integrity constraints, while checking constraint satisfaction using static analysis~\cite{Carbinetal2012,Carbinetal2013,chisel} or systematic testing~\cite{peforate1,hoffmann2011dynamic,zhao2023approxcaliper}. \Tool{} is an instance of such a general optimization framework for DNNs with certified robustness as its statistically-checked safety constraint.
For DNNs, researchers developed many testing and analysis techniques to assess the quality of DNN-based systems. Several papers proposed coverage and adequacy metrics~\citep{Ma_2018,Kim_2019,10.1145/3546947,neelofar2023reliableaiadequacymetrics}.
DeepHunter~\citep{10.1145/3293882.3330579} introduced coverage-guided fuzzing for DNNs. DeepMutation~\citep{ma2018deepmutationmutationtestingdeep} applied mutation testing to deep learning systems. \citet{10.1145/3377811.3380395} characterized faults in DNN-based systems, and DeepJanus~\citep{riccio2020modelbasedexplorationfrontierbehaviours} proposed search-based exploration of corner cases. 
While these complementary techniques detect and characterize failures in trained models, \Tool{} incorporates certified robustness directly into the quantization objective to produce compressed models {with strict statistical guarantees, offered by randomized smoothing.}
}

\vspace{-.05in}
\section{Conclusion}

We introduce \Tool{}, the first mixed-precision quantization framework that optimizes DNN accuracy and certified robustness under a computational budget. By using direct feedback from the quantized smoothed classifier's ACR, \Tool{} effectively searches for optimal quantization policies. Our experiments demonstrate that \Tool{} consistently outperforms state-of-the-art quantization methods, often reaching or improving FP32 accuracy and robustness using down to 0.84\% of operations. \Tool{} significantly reduces the computational requirements of randomized smoothing, enabling deployment in resource-constrained environments with certifiable robustness guarantees. Finally, \Tool{}'s design gives a blueprint for extending other quantization algorithms to support \mbox{certified robustness.}

\section*{Acknowledgments}
This research was supported in part by NSF Grants No. CCF-2217144, CCF-2313028, CCF-2238079, and CCF-2316233, a grant from the IBM-Illinois Discovery Accelerator Institute, a grant from the Amazon-Illinois Center on AI for Interactive Conversational Experiences (AICE), a Google Research Scholar Award, and an Open Philanthropy research grant.

\section*{Data Availability}

The source code and evaluation scripts for \Tool{} are available at \ArtifactURL.
\bibliographystyle{ACM-Reference-Format}
\bibliography{ref}

\end{document}